\documentclass[11pt]{article}

\usepackage[]{acl}

\usepackage{times}
\usepackage{latexsym}
\usepackage{graphicx}
\usepackage{booktabs}
\usepackage{adjustbox}
\usepackage{enumitem}
\usepackage{multirow}
\usepackage{tabularx}
\usepackage{tcolorbox}
\usepackage{listings}
\usepackage{amsmath}
\usepackage{arydshln}
\usepackage{makecell}

\usepackage[T1]{fontenc}
\usepackage[utf8]{inputenc}

\usepackage{microtype}

\usepackage{inconsolata}

\newcommand{\missingdetail}[1]{<missing\_detail> #1 </missing\_detail>}
\newcommand{\draftanswer}[1]{<draft\_answer> #1 </draft\_answer>}
\newcommand{\searchresult}[1]{<search\_result> #1 </search\_result>}

\definecolor{background_gray}{HTML}{FAFAFA}
\definecolor{frame_gray}{HTML}{424242}

\definecolor{background_blue}{HTML}{E3F2FD}
\definecolor{frame_blue}{HTML}{1565C0}

\definecolor{background_green}{HTML}{E8F5E9}
\definecolor{frame_green}{HTML}{2E7D32}

\definecolor{background_red}{HTML}{FFEBEE}
\definecolor{frame_red}{HTML}{C62828}

\definecolor{highlight_gray}{HTML}{E0E0E0}
\definecolor{highlight_yellow}{HTML}{FFF59D}

\title{Ask-before-Plan: Proactive Language Agents for Real-World Planning}

\author{Xuan Zhang$^1$,~ Yang Deng$^2\thanks{Corresponding author.}$,~ Zifeng Ren$^1$,~ See-Kiong Ng$^1$,~ Tat-Seng Chua$^1$\\
        $^1$National University of Singapore \quad $^2$Singapore Management University \\
    \texttt{\{xuanzhang,renzifeng\}@u.nus.edu} ~ \texttt{ydeng@smu.edu.sg} ~ \texttt{\{seekiong,dcscts\}@nus.edu.sg}}

\begin{document}
\maketitle
\begin{abstract}
The evolution of large language models (LLMs) has enhanced the planning capabilities of language agents in diverse real-world scenarios. Despite these advancements, the potential of LLM-powered agents to comprehend ambiguous user instructions for reasoning and decision-making is still under exploration. In this work, we introduce a new task, Proactive Agent Planning, which requires language agents to predict clarification needs based on user-agent conversation and agent-environment interaction, invoke external tools to collect valid information, and generate a plan to fulfill the user's demands. To study this practical problem, we establish a new benchmark dataset, Ask-before-Plan. To tackle the deficiency of LLMs in proactive planning, we propose a novel multi-agent framework, Clarification-Execution-Planning (\texttt{CEP}), which consists of three agents specialized in clarification, execution, and planning. We introduce the trajectory tuning scheme for the clarification agent and static execution agent, as well as the memory recollection mechanism for the dynamic execution agent. Extensive evaluations and comprehensive analyses conducted on the Ask-before-Plan dataset validate the effectiveness of our proposed framework.\footnote{The data and code will be released via \url{https://github.com/magicgh/Ask-before-Plan}.}
\end{abstract}

\section{Introduction}
\begin{figure}[t]
    \centering
    \setlength{\abovecaptionskip}{0pt}   
    \setlength{\belowcaptionskip}{0pt}
    \centering 
    \includegraphics[width=0.47\textwidth]{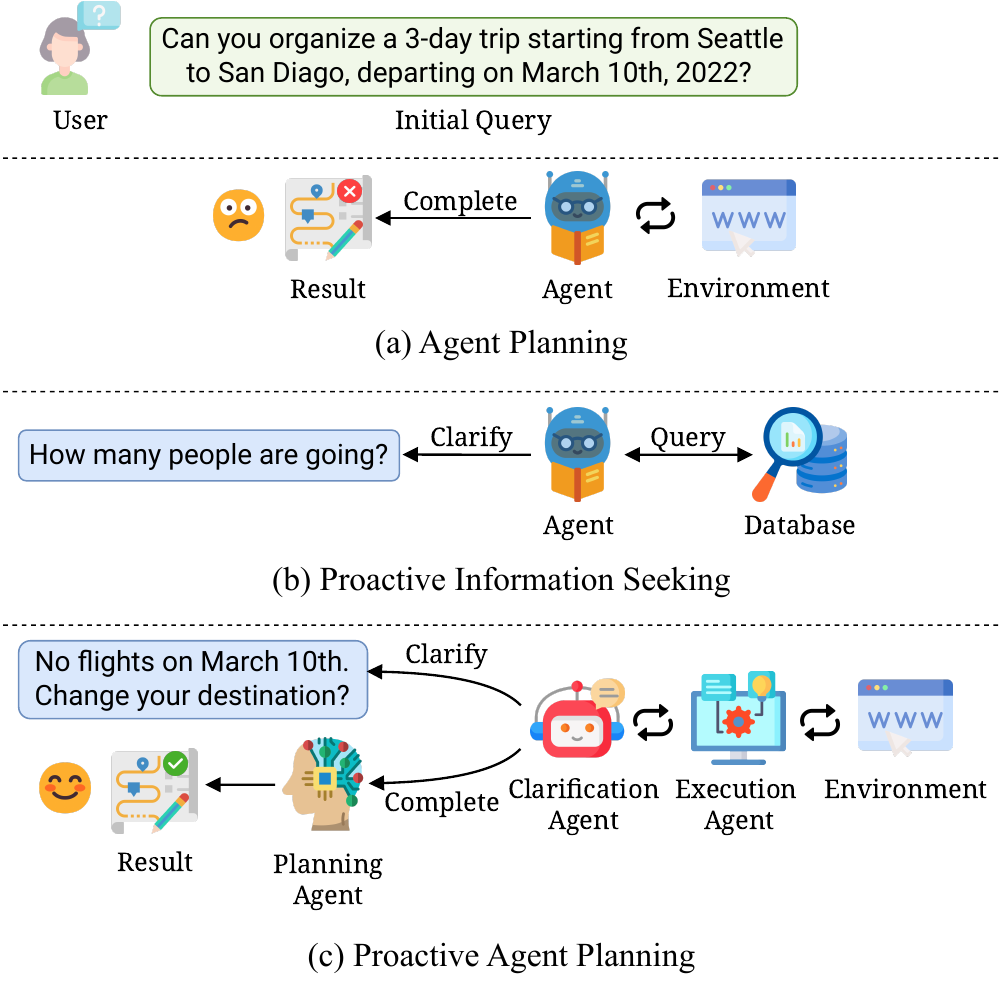}
    \caption{Illustrations of different problems.}
    \label{fig:task}
    \vspace{-0.45cm}
\end{figure}

Since long, people have been engaged in a never-ending trek to devise intelligent agents cloning appearance, behaviors, and even mindsets similar to human beings \citep{descartes,dreyfus}. 
The emergence of large language models (LLMs) and their integration into autonomous agents exhibits the potential for logical reasoning, decision-making, and problem-solving capabilities \citep{ruc-agent-survey,llm-agent-survey}. 
LLM-based language agents have been explored in various real-world planning problems, such as travel planning \cite{travelplanner}, web navigation \cite{mind2web}, online shopping \cite{webshop}, etc. 
Nevertheless, they struggle to \textit{digest} real-world instructions from time to time, particularly when the context is ambiguous or the user's intention is unclear. 
Figure \ref{fig:task}(a) exemplifies a case of \textit{Agent Planning}, where the language agent fails to produce a desired plan when encountering an unclear user instruction.

Inspired by proactive information-seeking systems \citep{proactive-dialogue-system,sigar23-proactive-conversational-agent} that can proactively ask clarification questions for clarifying the query uncertainty, latest studies on language agents \citep{tell-me-more,acl24-clamber} emphasize intention clarifications by incorporating implicit intention understanding into downstream applications (Figure \ref{fig:task}(b)). 
Despite their effectiveness in clarifying the uncertainty within user queries, there are several challenges that remain to be solved in agent planning scenarios: (1) The exclusive reliance on query understanding for clarification without considering the observations from the environment interaction; and (2) The negligence of the effect of user intention disambiguation in subsequent planning.

Given these important but neglected aspects, we formulate a novel task named \textbf{Proactive Agent Planning}. 
In addition to apprehending the user's implicit intentions in proactive information seeking, Proactive Agent Planning further requires language agents to predict clarification needs depending on the environmental observations, invoke external tools to collect valid information based on the user-agent conversation, and finalize decisions to accomplish the common goal conditioned on both user-agent conversation and agent-environment interaction history (Figure \ref{fig:task}(c)).
To study this practical problem, we first establish a new benchmark dataset, \textbf{Ask-before-Plan}, based on an agent planning dataset, TravelPlanner \citep{travelplanner}, which is used for evaluating the capability of agents for real-world travel planning.

To tackle the aforementioned challenges, we propose a novel multi-agent framework, named Clarification-Execution-Planning (\texttt{CEP}), to encompass user instruction clarification into downstream agent planning.
Specifically, the \texttt{CEP} involves three agents with distinct roles in \textit{Clarification}, \textit{Execution}, and \textit{Planning}. The \textit{Clarification} agent is responsible for understanding the uncertainty of user instructions and asking users clarifying questions to uncover their underlying intentions. 
The \textit{Execution} agent leverages various tools to interact with the environment, gathering necessary information for the \textit{Clarification} agent. 
The \textit{Planning} agent produces the final plan by aggregating the clarification process for accomplishing the initial user instruction. 
To supplement the deficiency of simply prompting LLMs to ask clarification questions or perform complex tool learning, we devise \textit{Trajectory Tuning} for fine-tuning the clarification and execution agents. 
Furthermore, we employ self-reflection to improve the reasoning process of the execution agent. However, the redundancy of self-reflection in multi-turn conversations may increase the time complexity of inference and introduce more noise into the context. To this end, we propose the memory recollection mechanism to optimize the memory utility for the execution agent in long-context reasoning.

In summary, our contributions are as follows:\vspace{-0.1cm}
\begin{itemize}[leftmargin=*]
    \item 
    We introduce a new and practical problem of \textbf{Proactive Agent Planning} to study the challenges of LLM-powered language agents in handling unclear user instructions.
    \item 
    We propose a novel multi-agent framework, namely \texttt{CEP}, which consists of clarification, execution, and planning agents, to address the underlying challenges in the Proactive Agent Planning problem. 
    \item 
    We construct the first dataset for studying Proactive Agent Planning, namely \textbf{Ask-before-Plan}. Extensive evaluations and comprehensive analyses in diverse settings validate the effectiveness of the proposed \texttt{CEP} framework.
\end{itemize}

\section{Related Work}

\paragraph{Language Agents}
Language-based agents \cite{llm-agent-survey,ruc-agent-survey,www24-agent-tutorial,sigir24-agent-tutorial} aim to perform real-world tasks that require professional expertise or extensive training by utilizing LLMs to conduct reasoning \cite{react,reflexion}, memory storage and retrieval \cite{dilu, memory-bank}, and tool use \cite{toolllm,toolformer}. 
The rapid development and application of language agents span various domains, including web agents \citep{mind2web, webshop, self-map}, game agents \citep{voyager, minecraft-ghost}, and medical agents \citep{hospital-agent, agent-clinic}, etc. 
Existing studies typically assume that the user instructions are clear enough for language agents to execute the task, while the instructions tend to be succinct and brief in actual scenarios, potentially leading to ambiguity and uncertainty.

\paragraph{Asking Clarification Questions}

The problem of asking clarification questions \cite{emnlp21-opendomain-clarification} typically involves two phases:
Clarification Need Prediction and Clarification Question Generation. 
Early studies develop ranking-based methods \citep{acl18-ranking, emnlp19-kbqa} to select clarification questions from a candidate pool, or apply end-to-end solutions \citep{emnlp22-pacific} for generating questions on the fly.
Recently, LLMs have also been employed to generate clarifying questions \citep{clam, procot, acl24-clamber, acl24-style}. Another line of research investigates environmental reward feedback to formulate questions \citep{www20-ir-clarification, clarify-delphi, star-gate}. \citet{tell-me-more} explore how language agents understand implicit user intentions but do not clarify users' needs based on their predefined preferences and consider the whole language agent problem, such as planning and task-solving.

%\paragraph{Task-oriented Dialogue} Task-oriented dialogue seeks to fulfill users' intentions and satisfy their needs. Traditional approaches \citep{acl18-sequicity, acl19-e3} pioneered the use of sequence-to-sequence (seq2seq) modeling with prior knowledge to track dialogue states. To address the diversity of user requirements resembling real-world situations, multi-domain dialogues were investigated \citep{multiwoz, aaai20-conversational-agent}. Furthermore, reinforcement learning has been used to enhance policy networks for knowledge base retrieval \citep{acl17-dialog-agent}, and function calling has been applied for dialogue state tracking \citep{dst-function}. Recently, the integration of task-oriented dialogue with LLM agents has improved decision-making capabilities \citep{decision-oriented-dialogue, chatshop}. Our work targets the combination of task-oriented dialogue with user interactions and environmental observations to enhance decision-making, particularly in long-horizon restrictive planning.

\begin{figure*}
    \setlength{\abovecaptionskip}{0pt}   
    \setlength{\belowcaptionskip}{0pt}
    \centering
    \includegraphics[width=\textwidth]{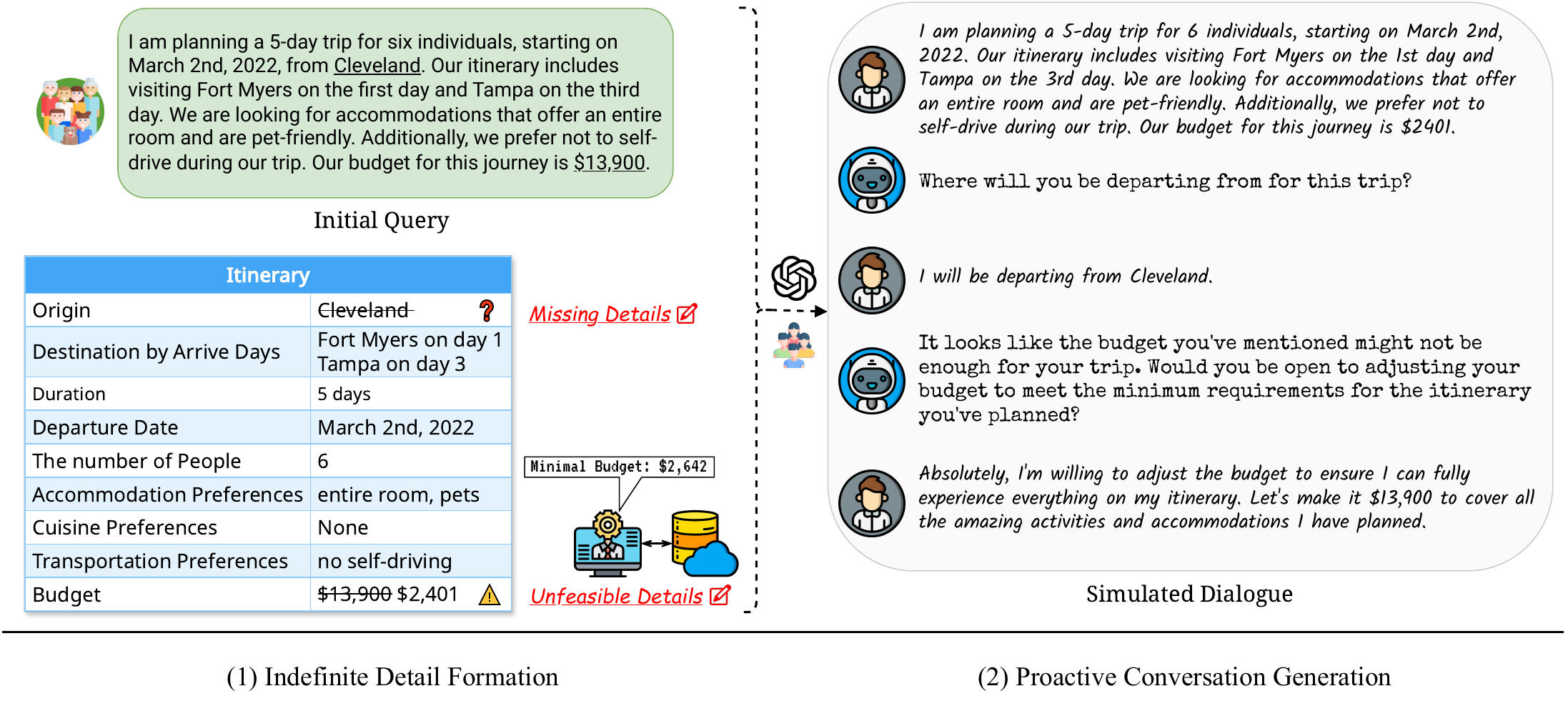}
    \caption{Data annotation process.}
    \label{fig:data}
    \vspace{-0.3cm}
\end{figure*}

\section{Ask-before-Plan Dataset}

Existing benchmarks on language agents either only focus on their capabilities of asking clarification questions, \textit{e.g.}, \textit{Tell Me More} \cite{tell-me-more} and \textit{Clamber} \cite{acl24-clamber}, or simply assume that all the user instructions are clear and explicit, \textit{e.g.}, \textit{TravelPlanner} \cite{travelplanner} and \textit{Mind2Web} \cite{mind2web}. 
In order to integrally investigate agents' competence in asking clarification questions as well as planning and interacting with the environment, we construct the \textbf{Ask-before-Plan} dataset. Built on the TravelPlanner benchmark, our dataset incorporates uncertain user instructions that require clarifications into real-world travel planning scenarios.

\subsection{Annotation \& Quality Control} 
%We specifically integrate proactive information-seeking dialogues 
We aim to incorporate uncertain user instructions with proactive information-seeking dialogues into tool-augmented planning, allowing the agent to ask clarification questions when faced with ambiguous or vague user instructions. As shown in Figure \ref{fig:data}, the dataset construction process is bifurcated into the following two phases: (1) \textbf{Indefinite Detail Formation}, and (2) \textbf{Proactive Conversation Generation}.

\paragraph{Indefinite Detail Formation} 
Each instruction in \textit{TravelPlanner} precisely describes the task and all pertinent details. To produce unclear instructions with indefinite details, we focus on two typical scenarios observed in proactive information-seeking dialogues: (1) clarifying under-specified queries, and (2) managing over-specified queries \cite{tacl2023-inscit,proactive-dialogue-system}:

\begin{enumerate}[nosep,leftmargin=*]
\item We create the under-specified user instruction with missing details based on attributes such as \textit{origin}, \textit{destination by arrival days}, \textit{departure date}, \textit{duration}, \textit{number of people}, and \textit{budget}, by discarding corresponding information from the original complete instruction.

\item 
%To construct over-specified instructions, we iterate through all possible values within the TravelPlanner dataset, selecting those that lack direct information matched in the environment. 
To construct over-specified instructions, we develop unfeasible details based on features such as \textit{accommodation}, \textit{cuisine}, and \textit{transportation} preferences, \textit{destination by arrival days}, and \textit{budget}. We exhaustively iterate over all possible values within the TravelPlanner environment to select those that lack directly matched information for instruction rewriting.

\end{enumerate}
Consequently, we range the size of indefinite details from 0 to 3 for each instruction to formulate the final dataset. Given the interdependencies among these details, topological sort is employed to reorder the details according to the dependency graph, and the priority of each detail is assigned accordingly. The descriptions of indefinite detail creation and instruction modification are presented in Appendix \ref{app:indefinite_detail}.

\paragraph{Proactive Conversation Generation}
After acquiring the revised instruction and selecting indefinite details, we utilize GPT-3.5 to simulate the proactive conversation between users and agents. In the user simulation, we adopt different tones, including succinct and passionate response patterns. To summarize, we follow the pipeline below to generate the real-world conversation:
%Utilizing the indefinite details identified in the initial phase, we employ \texttt{gpt-3.5-turbo-0125} to simulate dialogues between users and agents. Our simulation involves users adopting different tones, with a primary focus on both succinct and passionate responses. To efficiently construct the effective conversation to mimic the real-world conversation, we use the pipeline %including Initial Query Proposal, Target Detail Grounding (Observation Grounding), Guided Question Generation, Draft Answer Confinement, User Response Generation, and Human Examination shown as follows:
%\begin{enumerate}[leftmargin=*]
%    \item Initial query converted from JSON object.
 %   \item Observation prompts for missing or unfeasible details.
 %   \item Question generated based on initial query and observation with examples.
  %  \item Draft answer as thoughts to prompt GPT-3.5 to simulate the user.
 %   \item User response generated based on the question and draft answer.
  %  \item Human examination after the completion of GPT-3.5 generation. We apply the rule-based method to check whether the details mentioned in the question and keywords of draft answers mentioned in the generated answer. If any of the generated content is not passed, we will manually modify to make sure the correctness. 
%\end{enumerate}
\begin{enumerate}[nosep,leftmargin=*]
        \item \textbf{Ambiguity Observation}: The description of an unclarified indefinite detail is presented to GPT-3.5 (Agent) as the observation.
        \item \textbf{Question Generation}: A question is generated to inquire about the indefinite detail based on the initial instruction and observation.
        \item \textbf{Draft Answer}: The draft answer is provided to GPT-3.5 (User) to guide the clarification of the indefinite detail.
        \item \textbf{Natural Response Generation}: A natural user response is generated based on the question and the draft answer.
        \item \textbf{Quality Control}: The human examiner verifies that the question covers the indefinite detail and that the response provides clarification.
\end{enumerate}
The detailed implementation of these simulations, including the prompts used, is documented in Appendix \ref{app:proactive_conversation}. The methodology achieved a pass rate of 87.2\% on the human examination. %without human interference. \dy{what is the success rate? what will be done if it fails?}

\subsection{Dataset Statistics}
After the dataset construction, we obtain 1,000 samples for both the training and testing datasets. Indefinite details were randomly selected from 11 distinct candidate types, as previously defined. The distribution of indefinite detail sizes is structured as follows: 10\% of the data contains no indefinite details, and 30\% of the data includes 1, 2, or 3 indefinite details each. The overall dataset statistics are presented in Table \ref{tab:data_statistics}. 
The environment setting is presented in Appendix \ref{app:environment}. 

\begin{table}[t]
\scriptsize
\setlength{\abovecaptionskip}{5pt}   
\setlength{\belowcaptionskip}{0pt}
\centering
\begin{tabular}{p{1.9cm}lrr}
\toprule
                       &      & Train & Test \\ \midrule
\multirow{3}{*}{\textbf{Level}}& Easy                         & 331   & 348  \\
& Medium                       & 336   & 333  \\
& Hard                         & 333   & 319  \\ \midrule
\multirow{3}{*}{\textbf{Duration}} &  3 days                       & 335   & 308  \\
& 5 days                       & 337   & 351  \\
& 7 days                       & 328   & 341  \\ \midrule
\multirow{7}{*}{\textbf{Missing Details}} 
& Origin                       & 219   & 192  \\
& Destination by Arrival Days & 379   & 360  \\
& The number of People         & 200   & 204  \\
& Duration                     & 199   & 201  \\
& Departure Date               & 199   & 213  \\
& Budget                       & 384   & 401  \\ 
& \textbf{Total}     & 1,204  & 1,183 \\
\midrule
\multirow{6}{*}{\textbf{Unfeasible Details}} 
& Destination by Arrival Days & 189   & 176  \\
& Accommodation Preferences    & 182   & 185  \\
& Cuisine Preferences          & 18    & 20   \\
& Transportation Preferences   & 20    & 24   \\
& Budget                       & 187   & 212  \\ 
& \textbf{Total}  & 596   & 617  \\
\midrule
\multirow{2}{*}{\textbf{Total}} & \# Dialogues & 1,000 & 1,000\\
& \# Turns & 2,800 & 2,800 \\
%Avg. \# Turns / Dialogue & 2.8 & 2.8\\
%Avg. Question Length & 167.60 & 113.42\\
%Avg. Answer Length & 168.06 & 121.10\\
\bottomrule
\end{tabular}
\caption{Dataset statistics of \textbf{Ask-before-Plan}.}
\label{tab:data_statistics}
\vspace{-0.3cm}
\end{table}

\subsection{Problem Definition}\label{sec:problem_def}
Regarding the problem of \textbf{Proactive Agent Planning}, given an initial user instruction $q_0$ potentially containing $T$ vague or unfeasible details, $D = \{d_1, \ldots, d_T\}$, we define three subtasks:
\begin{itemize}[leftmargin=*]
    \item \textbf{Clarification Need Prediction \& Clarification Question Generation}: For a user-agent dialogue up to $t-1$ turns, we define the conversation history, $C_{t-1} = \{(q_0), (a_1, q_1), \ldots, (a_{t-1}, q_{t-1})\}$, the agent-environment interaction history based on $C_{t-1}$, denoted as $E_{t-1} = \{f_1, o_1, \ldots, f_n, o_n\}$, where $f_i$ is the function call and $o_i$ is the observation from the environment, and the unclarified detail set for this turn, $D_t = \{d_{t+1}, \ldots, d_T\}$. The clarification agent is tasked with predicting a binary label $b_t$, indicating the need for clarification based on the vagueness and feasibility of environmental responses $E_{t-1}$, and eliciting unclarified details in $D_t$ by generating a clarification question $a_t$ if needed. The conversation will advance according to the topological priority assigned to each detail, with those sharing the same priority addressed randomly. %\zx{When it comes to the final conversation turn $T$, the agent is expected to predict $C_T$ as clear, and no clarified questions should be generated.}
    \item \textbf{Tool Learning}: Given the ongoing conversation  $C_{t} = \{(q_0), (a_1, q_1), \ldots, (a_{t-1}, q_{t-1}), a_t\}$, the agent needs to generate the complete interaction chain $E_{t}^{i-1} = \{f_1, o_1, \ldots, f_{i-1}, o_{i-1}\}$ within the present information. A correct function call is defined as having accurate function names and valid parameters. This stage is used to interact with the environment as well as provide observation for the next-turn clarification.
    \item \textbf{Agent Planning}: Upon completion of the clarified dialogue $C_t$ and the last interaction history $E_t$, the planning agent must generate a valid plan in JSON format with pre-defined fields such as \textit{day}, \textit{current city}, \textit{transportation}, etc. %, \textit{accommodation} "breakfast", "lunch", "dinner", "accommodation", "attraction".
\end{itemize}
\section{Method}

We introduce our multi-agent framework, namely Clarification-Execution-Planning (\texttt{CEP}), which consists of three agents: clarification agent, execution agent, and planning agent.

%For the task of interacting with environment using tools, we proposed two settings, namely static interaction and dynamic interaction.  
Specifically, we investigate two settings of execution agents to interact with the environment using tools, namely static and dynamic interaction.
For the static interaction, the agent needs to generate the whole valid interaction of tool calls $E_t^{i-1}$ based on the current conversation $C_t$ in one-step inference. We regard the dynamic interaction as a multi-step inference. In this situation, the agent is expected to generate one action $f_i$ depending on the current conversation $C_t$, with previous interaction history $E_t^{i-1} = \{f_1, o_1, \ldots, f_{i-1}, o_{i-1} \}$.
% Naming of each module? ask: clarification? asking questions? consultation? tool: navigation? interaction? tool use? execution? plan: plan
% Naming of details: missing/unfeasible?
%In this paper, we proposed the trajectory tuning for ask and static tool-use task. In order to explore the setting of dynamic tool-use, we introduce the shared memory module into the agent.

\begin{figure}
    \setlength{\abovecaptionskip}{5pt}   
    \setlength{\belowcaptionskip}{0pt}
    \centering
    \includegraphics[width=0.48\textwidth]{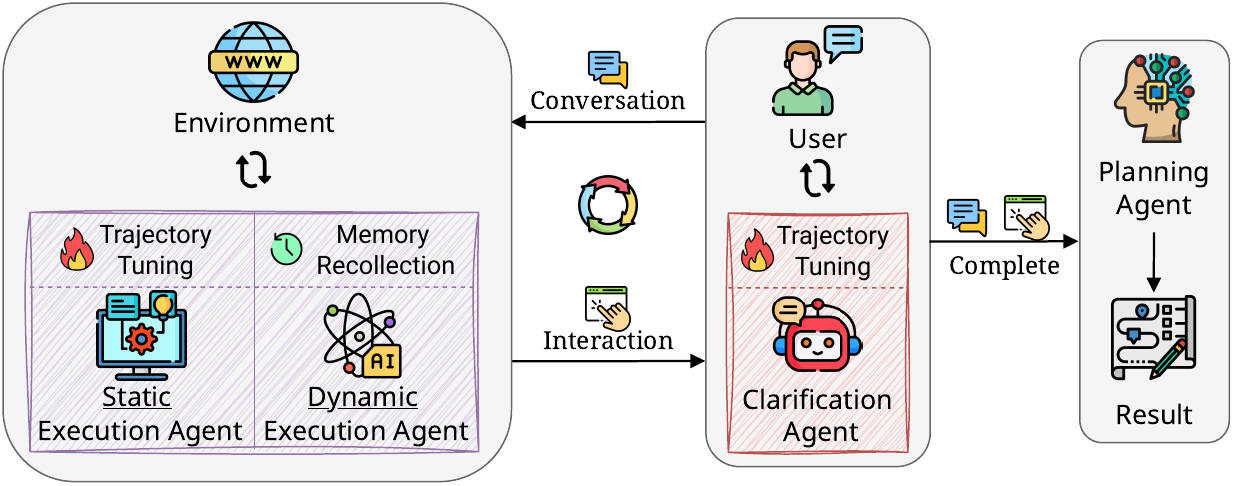}
    \caption{Overview of the \texttt{CEP} framework.}
    \label{fig:framework}
    \vspace{-0.3cm}
\end{figure}

As illustrated in Figure \ref{fig:framework}, we introduce the trajectory tuning scheme for the clarification agent and static execution agent, along with memory recollection for the dynamic execution agent. Finally, the planning agent generates a valid plan based on the conversation history and interaction results.

\subsection{Trajectory Tuning}
Concerning the status quo, LLMs struggle to spontaneously clarify users' intentions during the conversation \citep{proactive-dialogue-system} and accurately obtain necessary information via tool utilization without hallucination \citep{formal-llm}, even after the instruction tuning or providing the well-crafted prompts. To resolve these issues and align the current language-based agents with our task, we propose to use the trajectory $(C_{t-1}, E_{t-1})$ to finetune the clarification agent and the execution agent in the static setting. We sample $t-1$ details according to topological order (see Appendix \ref{app:indefinite_detail}) to craft $C_{t-1}$ and derive the valid interaction chain $E_{t-1}$ from $C_{t-1}$ based on ground truth. 

\paragraph{Clarification Agent}
As introduced in Section \ref{sec:problem_def}, there are two steps to be achieved for the clarification agent:
(1) Clarification Need Prediction. In this step, the agent generates a boolean indicator $b_t$ to predict the need for clarification based on the ongoing conversation $C_{t-1}$ and the last turn interaction $E_{t-1}$. (2) Clarification Question Generation. If $b_t$ is true, the model needs to generate a clarifying question $a_t$ for a specific detail $d_t$. In this case, we create a conversation sequence $[f_{\mathrm{prompt}}(C_{t-1}, E_{t-1}, b_t), {a_t}^{[b_t = 1]}]$ and tokenize it into the input tokens $\mathbf{x}_t$, where $f_{\mathrm{prompt}}(\cdot)$ is a function to apply the prompt template.

\paragraph{Static Execution Agent}
In the tool learning subtask, we apply trajectory tuning for the static setting. Given the current-turn conversation $C_{t}$ and the complete interaction chain $E_{t}$, we construct the sequence $[f_{\mathrm{prompt}}(C_{t}), E_{t}]$ and transform it into the input $\mathbf{x}_t$.

\paragraph{Training Objective}
Finally, we train the agent using the trajectory sequence $\mathbf{x}_t$ for turn $t$ in the autoregressive manner: 
\begin{equation}
    \mathcal{L}=\max_{\theta}\frac{1}{|T|}\sum\nolimits_{t=1}^T \sum\nolimits_{i=1}^{N_t}\log P_{\theta}^t(x^t_i|\mathbf{x}^t_{<i})
\end{equation}
where $\theta$ denotes the model parameter to be learned, $T$ denotes the number of total conversation turns, and $N_t$ denotes the number of tokens in turn $t$.

\subsection{Memory Recollection}
In view of dynamic tool interaction, Reflexion \cite{reflexion} has been validated as an effective approach to improve the reasoning capability of LLMs by incorporating the self-reflective feedback into the ReAct framework \cite{react}. However, when applying this approach to Proactive Agent Planning, there is a potential issue: the dynamic execution agent may encounter similar types of exceptions and repeatedly generate identical rationales across different turns.
To alleviate this disadvantage, we introduce the \textbf {Memory Recollection} mechanism to \textit{reuse} self-reflective feedback from previous turns.

%For tool learning in the dynamic setting, we introduce the memory recollection. Inspired by Reflexion \cite{reflexion}, which incorporates the short-term memory and the long-term memory into the ReAct framework \cite{react}. Considering the nature of the conversation in our task is to request for new information as the number of conversational turns increases. For each conversation turn $C_t$, the execution agent will generate all function calls and get the response from the environment. 

% To apply Reflexion into multi-turn conversations, given the context $C_t$, the model will generate $f_i$ based on the interaction history $E_t^{i-1}$, where $t$ denotes $t$-th turn in the conversation, and $i$ denotes $i$-th iteration in the interaction. Memory recollection mechanism maintains a long-term working memory $R_t^i$ for self-reflection. Specifically, we prompt the model to generate reflective text $r_i$ if $f_i$ is failed, and append $r_i$ into $R_t^i$:
Given the conversation context $C_t$ at turn $t$, the execution agent generates a tool call $f_i$ during the $i$-th interaction, based on previous interactions $E_t^{i-1}$ and all reflective text in the memory bank $R_t^i$, \textit{i.e.}, $f_i = \text{LLM}(C_t, E_t^{i-1}, R_t^i)$. When $f_i$ is invalid, the agent will generate a rationale $r_i$ and store it into $R_t^{i+1}$ for self-reflection at the next interaction:
\begin{equation}
    R_t^{i+1} =
    \begin{cases} 
      R_t^{i} & \text{if } f_i \text{ is valid} \\
      \text{concat}(R_t^{i}, r_i) & \text{else}
    \end{cases}
\end{equation}

% Given that different conversation turns to accomplish the same user instruction share the common information, resulting in the repetitive function calls among each turns, the memory recollection can accumulate the previous self-reflective feedback for subsequent turns. Formally, the reflexion memory bank for turn $t+1$ inherits from turn $t$, \textit{i.e.}, $R_{t+1} = R_t$. This mechanism not only prevents the model from repeating mistakes made in previous turns, but also reduces the inference time. The theoretical time complexity analysis and corresponding experimental results are presented in the Appendix \ref{app:time_complexity}. 
Given that $C_t$ shares common clarified details across different turns, the memory recollection accumulates the reflective feedback from previous turns, \textit{i.e.}, $R_{t+1} = R_t$. This mechanism not only prevents the model from repeating mistakes in history turns but also reduces the inference time for upcoming turns. The theoretical time complexity analysis and corresponding experimental results are presented in Appendix \ref{app:time_complexity}.

\subsection{Overall Framework}
Given the conversation $C_t$ of turn $t$, the execution agent first generates the whole interaction chain $E_t$. %In the dynamic setting, the interaction chain is generated step by step. 
Next, based on $C_t$ and $E_t$, the clarification agent is required to predict the clarification need and ask a question about a detail from $D_t$ if necessary. %In the last conversation turn $T$, 
This process continues iteratively until all indefinite details have been recovered by the clarification agent.
Once all such details are clarified, %and based on the conversation $C_T$ and the interaction $E_T$ from $C_T$, 
the planner agent will generate a valid plan in the format of JSON, based on the conversation $C_T$ and the interaction $E_T$ from $C_T$. %Formally, a JSON plan is generated by a planning agent, \textit{i.e.}, $\textrm{Planner}(C_T, E_T)$.

\section{Experiment}
\subsection{Experimental Setups} 
\paragraph{Baselines}
%Since we define three independent subtasks, for the tool and plan subtasks, we adapt traditional methods as our baselines. For the asking agent, because there does not exist the proactive agent that both consider the environment and conversation together, thus we devise the environment-only method, which is the rule-based information-seeking agent, and the conversation-only method. For conversation-only method, we choose Proactive and Proactive CoT (ProCoT) \cite{procot}. For the tool agent, we incorporate the brute force method (which is to call all the tools whether it's valid or not), direct, ReAct \cite{react}, and Reflexion \cite{reflexion}. For the planning agent, our baselines include  Direct, Zero-shot CoT \cite{zero-shot-cot}, ReAct \cite{react}, and Reflexion \cite{reflexion}.
%\dy{In terms of the three subtasks in the problem of proactive agent planning, we adopt corresponding baselines as follows: 
We adopt the following baselines for subtasks in the problem of Proactive Agent Planning: 
% (1) For the clarification subtask, we adopt one rule-based environment-only method and two conversation-only methods as baselines, including Proactive and Proactive CoT (ProCoT) \cite{procot}. Since there is no existing approach that concurrently consider the environment and conversation for asking clarification questions, we further include the direct zero-shot GPT-3.5 with the same prompts as \texttt{CEP}. 
% (2) In tool learning subtask, we adopt four baseline methods, including brute-force, direct zero-shot prompting, ReAct \cite{react}, and Reflexion \cite{reflexion}. 
% (3) For the planning subtask, our baselines include  Direct Zero-shot, Zero-shot CoT \cite{zero-shot-cot}, ReAct \cite{react}, and Reflexion \cite{reflexion}. 
(1) Clarification: We first design a rule-based environment-only method and adopt four conversation-only methods, including Proactive, Proactive CoT (ProCoT) \cite{procot}, and Direct prompting, as baselines. For the reason that there is no existing approach that concurrently considers the environment and conversation when asking clarification questions, we further directly prompt the GPT-3.5 with the conversation history and interaction log, along with In-context Learning (ICL) prompting, as additional baselines. (2) Tool learning: We adopt the general brute-force algorithm and four baselines, including Direct prompting (GPT-3.5) and finetuned ToolLLM (LLaMA-2-7B) \cite{toolllm} for the static execution setting, as well as ReAct \cite{react} and Reflexion \cite{reflexion} for the dynamic setting. 
(3) Planning: Our baselines include Direct prompting, Zero-shot CoT \cite{zero-shot-cot} for static settings, as well as ReAct \cite{react} and Reflexion \cite{reflexion} for dynamic settings.\footnote{ReAct and Reflexion are not tested for Mistral-7B due to the inability to generate valid JSON plans.} Additionally, Greedy Search is included as a baseline for comparison. 
Details of these baselines are presented in Appendix \ref{app:baseline}.

\paragraph{Evaluation Metrics}
Similarly, we employ evaluation metrics for the three subtasks as follows: 
(1) In the clarification subtask, Clarification Accuracy (Clarif. Acc) is adopted for evaluating the performance of Clarification Need Prediction, and meanwhile, we setup three metrics to score the quality of clarification question generation, including Rule-based, BLEU, and GPT scores. 
(2) For the tool learning subtask, following previous studies \cite{tool-ste}, we also adopt the same metrics for evaluation, including Well-formed, API Match, and Correctness. Furthermore, we add two metrics to assess the Repeat Rate and Pass Rate of tool use. 
(3) Concerning the planning subtask, we also adopt the same evaluation metrics as TravelPlanner \cite{travelplanner}, including Delivery Rate, Commonsense Pass Rate, Hard Constraint Pass Rate, and Final Pass Rate. 
Details of these evaluation metrics are presented in Appendix \ref{app:metric}.

\paragraph{Implementation Details}
For all experiments involving GPT-3.5, we use \texttt{gpt-3.5-turbo-0125}. Mistral-7B, LLaMA-2-7B, and LLaMA-3-8B are adopted for finetuning and inference. More implementation details are presented in Appendix \ref{app:implementation}. 

\begin{table}[t]
    \setlength{\abovecaptionskip}{5pt}   
    \setlength{\belowcaptionskip}{0pt}
    \centering
    \setlength{\tabcolsep}{1mm}{
    \begin{adjustbox}{max width=0.48\textwidth}
    \begin{tabular}{lccccccc}
\toprule
 &
  \multicolumn{2}{c}{Clarif. Acc} &
  \multicolumn{2}{c}{Rule-based Score} &
  \multicolumn{1}{c}{} &
  \multicolumn{2}{c}{GPT Score} \\ \cmidrule(lr){2-3} \cmidrule(lr){4-5} \cmidrule(lr){7-8} 
 &
  \multicolumn{1}{c}{Micro} &
  \multicolumn{1}{c}{Macro} &
  \multicolumn{1}{c}{Micro} &
  \multicolumn{1}{c}{Macro} &
  \multicolumn{1}{c}{\multirow{-2}{*}{BLEU}} &
  \multicolumn{1}{c}{Micro} &
  \multicolumn{1}{c}{Macro} \\ \midrule
Environment-only      & 70.4 & 17.7 & 21.5 & 8.1 & 1.0 & 40.1 & 19.4 \\ \midrule
\multicolumn{8}{c}{\cellcolor{highlight_gray}\textbf{Conversation-only}}          \\ \midrule
Proactive (GPT-3.5)            & 62.3 & 6.1 & 9.7 & 3.4 & 3.7 & 0.9 & 0 \\
ProCoT (GPT-3.5)                & 33.7 & 10.6 & 3.3 & 2.6 & 2.2 & 2.4 & 1.8 \\
Direct (Mistral-7B) & 59.4 & 24.6 & 56.8 & 50.8	& 47.9 & 65.8 & 59.3 \\

Direct (LLaMA-3-8B) & \textbf{76.8} & \textbf{48.5} & \textbf{70.5} & \textbf{64.6}	& \textbf{53.4} & \textbf{80.7} & \textbf{75.5} \\
\midrule
\multicolumn{8}{c}{\cellcolor{highlight_gray}\textbf{Environment + Conversation}} \\ \midrule
Direct (GPT-3.5)            & 47.0 & 16.9 & 20.8 & 17.4 & 8.2 & 8.6 & 6.2 \\
ICL (GPT-3.5) & 65.7 & 29.4	& 2.1 & 0.6 & 8.8 & 2.7 & 0.9 \\
\texttt{CEP} (Mistral-7B)        & 82.8 & 51.7 & 54.2 & 37.0 & 44.5 & 73.1 & 58.6 \\ 
\texttt{CEP} (LLaMA-3-8B)            & \textbf{99.4} & \textbf{98.2} & \textbf{69.7} & \textbf{55.8} & \textbf{57.2} & \textbf{85.8} & \textbf{77.0} \\
\bottomrule
\end{tabular}
    \end{adjustbox}}
    \caption{Evaluation of clarification.}
    \label{tab:ask_overall}
    \vspace{-0.3cm}
\end{table}

\subsection{Overall Evaluation}
We list our experiments on clarification, tool learning, and planning subtasks, shown in Tables \ref{tab:ask_overall}, \ref{tab:tool_overall}, \ref{tab:plan_overall}. 
To assess the capability of agents on these three subtasks individually, we use the ground truth of prior subtasks as input for the subsequent subtask. 
Evaluations of the integral framework are discussed in Section \ref{sec:ablation_study}. 
%In light of the further exploration of adopting the generated response as input for the subsequent subtask, please refer to Section \ref{sec:ablation_study}. 

\subsubsection{Evaluation on Clarification}
%In clarification task, we regard the GPT Score as the main metric to evaluate the quality of generated questions. We also provide two auxiliary metrics for reference. 
From Table \ref{tab:ask_overall}, the results of the environment-only method surpass Proactive and ProCoT, pointing out the importance of using environment observation as a supervised signal in clarification for language agents. 

Comparing four conversation-only methods, Proactive performs well in turn-grained (Micro Clarif. Acc), while ProCoT strengthens in instance-grained clarification (Macro Clarif. Acc). 
Upon careful inspection, we observe that Proactive prefers to clarify the user's instruction, whereas ProCoT tends to directly execute the instruction, which is discussed in Section \ref{sec:detailed_analysis}. Besides, ProCoT performs better in clarification question generation with higher GPT scores. Both Direct prompting methods deliver better results than the previous two baselines while achieving BLEU and GPT scores that are closer to their Environment + Conversation counterparts. However, they fail to distinguish the need for clarifying the conversation, highlighting the importance of observing both environment and conversation.

For approaches using both environment and conversation observations, Direct (GPT-3.5) surpasses Proactive and ProCoT in question quality and macro clarification accuracy. However, it still performs worse than the environment-only method, regardless of Clarification Need Prediction or Clarification Question Generation. %, probably because we use the zero-shot setting. 
%It's not surprising that 
ICL (GPT-3.5) performs better in clarification need prediction but worse in question generation, supporting the argument that in-context learning is an unstable approach for the clarification subtask.
Finally, our proposed \texttt{CEP} framework with either Mistral-7B or LLaMA-3-8B outperforms all other methods. Specifically, finetuned LLaMA-3-8B is better than finetuned Mistral-7B, mainly due to the larger parameter size and better pretraining data. %Lastly, we found the two auxiliary evaluation metrics we present, Rule-based Score and BLEU, approximately align with our main metric, GPT Score.

\begin{table}[t]
    \setlength{\abovecaptionskip}{5pt}   
    \setlength{\belowcaptionskip}{0pt}
    \centering
    \setlength{\tabcolsep}{1mm}{
    \begin{adjustbox}{max width=0.48\textwidth}
    \begin{tabular}{lcccccccc}
\toprule
 &
  \multicolumn{1}{c}{\multirow{2}{*}{Well-formed}} &
  \multicolumn{1}{c}{\multirow{2}{*}{\makecell{API\\Match}}} &
  \multicolumn{1}{c}{\multirow{2}{*}{\makecell{Repeat\\Rate}}} &
  \multicolumn{3}{c}{Correctness} &
  \multicolumn{2}{c}{Pass Rate} \\ \cmidrule(lr){5-7} \cmidrule(lr){8-9}
 &
  \multicolumn{1}{c}{} &
  \multicolumn{1}{c}{} &
  \multicolumn{1}{c}{} &
  \multicolumn{1}{c}{P} &
  \multicolumn{1}{c}{R} &
  \multicolumn{1}{c}{F1} &
  \multicolumn{1}{c}{Micro} &
  Macro \\ \midrule
Brute-force    & 100 & 98.8 & 0 & 77.2 & 90.3 & 81.9 & 45.7 & 22.0 \\ \midrule
\multicolumn{9}{c}{\cellcolor{highlight_gray}\textbf{Static Setting}}  \\ \midrule
Direct (GPT-3.5)  & 99.9 & 88.9 & 0.07 & 72.8 & 62.0 & 64.7 & 7.1 & 2.3\\ 
ToolLLM (LLaMA-2-7B)	& 99.7 & 82.9 & 2.4	& 65.9 & 66.1 & 63.3 & 16.3 & 3.4 \\
\texttt{CEP} (Mistral-7B) & 99.4 & 93.4 & 0.15 & 91.7 & 90.1 & 90.1 & 57.6 & 27.3\\
\texttt{CEP} (LLaMA-3-8B)     & \textbf{100} & \textbf{99.3} & \textbf{0.04} & \textbf{97.9} & \textbf{98.1} & \textbf{97.9} & \textbf{89.0} & \textbf{78.4} \\ \midrule
\multicolumn{9}{c}{\cellcolor{highlight_gray}\textbf{Dynamic Setting}} \\ \midrule
ReAct (GPT-3.5)   & 66.2 & 33.3 & 14.3 & 42.8 & 15.6 & 21.1 & \textbf{1.4} & 0 \\
Reflexion (GPT-3.5)  &  70.5 & 42.2 & 11.1 & 44.4 & 18.6 & 24.0 & 1.0 & 0 \\
\texttt{CEP} (GPT-3.5) & \textbf{73.3} & 45.3 & \textbf{9.8} & 45.0 & 19.2 & 24.7 & 1.1 & 0 \\
\cdashline{1-9} \noalign{\vskip 0.75ex}
ReAct (Mistral-7B)  &  49.0 & \textbf{50.0} & 11.8 & \textbf{58.1} &\textbf{24.6} & \textbf{32.3} & 1.2 & 0 \\
Reflexion (Mistral-7B) & 48.1 & 46.3 & 15.6 & 54.2 & 21.8 & 28.8 & 1.2 & 0 \\
\texttt{CEP} (Mistral-7B) & 46.9 & 42.4 & 18.1 & 49.0 & 19.5 & 25.6 & 1.3 & 0 \\ \bottomrule
\end{tabular}
    \end{adjustbox}}
    \caption{Evaluation of tool learning. }
    \label{tab:tool_overall}
    \vspace{-0.3cm}
\end{table}

\subsubsection{Evaluation on Tool Learning}
As shown in Table \ref{tab:tool_overall}, we evaluate tool learning in two settings, namely static and dynamic. 

\paragraph{Discussion on Static vs. Dynamic Settings}
First of all, Brute-force outperforms all methods in the dynamic setting. In general, agents in the dynamic setting represent worse results compared with agents in the static setting, indicating that in a complex environment similar to the real-world situation, agents are more likely to fail to handle different exceptions during the interaction. 

\paragraph{Comparison on Static Setting}
In the static setting, we observe the same phenomenon in the clarification subtask, \textit{i.e.}, LLaMA-3-8B performs better, and even surpasses the Brute-force baseline. Zero-shot GPT-3.5 presents the undesired performance on API match, correctness, and pass rate. Interestingly, ToolLLM also produces unexpected results, suggesting that the pretraining knowledge from ToolBench did not translate into improved performance on our task.

%, but still better than all dynamic agents. 

\paragraph{Comparison on Dynamic Setting}
In the dynamic setting, due to the input token limitation of LLaMA-3-8B, we could only conduct experiments over GPT-3.5 and Mistral-7B for three different strategies. We find that GPT-3.5 is prone to generating actions without syntax errors. But when it comes to evaluating the quality of the generated actions, Mistral performs better. Secondly, we find the performance degradation trend of Mistral from the strategy \textit{ReAct} to our proposed memory recollection mechanism, which is in contrast to the observation we find in GPT-3.5, probably because of the reasoning ability drops when processing the context with length over 8K. The results of GPT-3.5 prove the effectiveness of memory recollection. It can also be concluded that in a dynamic setting, all agents fail to consistently generate correct actions throughout the entire interaction. More analysis on dynamic execution agents is presented in Appendix \ref{app:dynamic_execution}. 

\subsubsection{Evaluation on Planning}
In the planning task, greedy search achieves a delivery rate of 100\% and meets most constraints, but this does not result in a valid plan. For Mistral, we only test the Direct and CoT strategies because the agent gets trapped in a fail-retry loop when using ReAct and Reflexion. This scenario mirrors the performance drop observed in the tool learning task with a super-long context. For GPT-3.5, we observe the same trend in TravelPlanner \cite{travelplanner}, the performance degrades from direct prompting to Reflexion, demonstrating the difficulty for most advanced agents to finish the planning task. Among these results, only GPT-3.5 with direct prompting successfully generates one passed plan out of 1,000 samples in the test set, indicating the significant challenge of this problem.

\begin{table}[t]
    \setlength{\abovecaptionskip}{5pt}   
    \setlength{\belowcaptionskip}{0pt}
    \centering
    \setlength{\tabcolsep}{1mm}{
    \begin{adjustbox}{max width=0.48\textwidth}
    \begin{tabular}{lcccccc}
\toprule
 & & \multicolumn{2}{c}{Commonsense}& \multicolumn{2}{c}{Hard Constraint} \\
 &
 \multicolumn{1}{c}{\multirow{2}{*}{\makecell{Delivery\\Rate}}} &
  \multicolumn{2}{c}{\begin{tabular}[c]{@{}c@{}}Pass Rate\end{tabular}} &
  \multicolumn{2}{c}{\begin{tabular}[c]{@{}c@{}}  Pass Rate\end{tabular}} &
  \multicolumn{1}{c}{\multirow{2}{*}{\makecell{Final\\Pass Rate}}} \\ \cmidrule(lr){3-4} \cmidrule(lr){5-6} 
 &
  \multicolumn{1}{c}{} &
  \multicolumn{1}{c}{Micro} &
  \multicolumn{1}{c}{Macro} &
  \multicolumn{1}{c}{Micro} &
  \multicolumn{1}{c}{Macro} &
  \multicolumn{1}{c}{} \\ \midrule
Greedy Search & 100 & 76.9 & 0 & 64.5 & 46.7 & 0 \\ 
\cdashline{1-7} \noalign{\vskip 0.75ex}
Direct (Mistral-7B) & 86.6 & 44.8 & 0.4 & 4.0 & 0.9 & 0 \\
CoT (Mistral-7B)  & 61.5 & 29.8 & 0 & 2.4 & 0.1 & 0 \\
\cdashline{1-7} \noalign{\vskip 0.75ex}
Direct (GPT-3.5) & \textbf{98.6} & \textbf{63.7} & \textbf{0.7} & \textbf{19.0} & \textbf{5.1} & \textbf{0.1} \\
CoT (GPT-3.5)  & 77.5 & 50.0 & 0.6 & 16.2 & 5.2 & 0 \\
ReAct (GPT-3.5)  & 68.7 & 38.0 & 0 & 3.2 & 0.6 & 0 \\
Reflexion (GPT-3.5) & 61.5 & 33.9 & 0 & 3.1 & 0.4 & 0 \\ \bottomrule
\end{tabular}
    \end{adjustbox}}
    \caption{Evaluation of planning.}
    \label{tab:plan_overall}
    \vspace{-0.1cm}
\end{table}
\begin{table}[t]
    \setlength{\abovecaptionskip}{5pt}   
    \setlength{\belowcaptionskip}{0pt}
    \centering
    \setlength{\tabcolsep}{1mm}{
    \begin{adjustbox}{max width=0.48\textwidth}
\begin{tabular}{lcccccc}
\toprule
\multicolumn{7}{c}{\cellcolor{highlight_gray}\textbf{Clarification}} \\ \midrule
 &
  \multicolumn{2}{c}{Clarif. Acc} &
  \multicolumn{2}{c}{Rule-based Score} &
  \multicolumn{2}{c}{GPT Score} \\ \cmidrule(lr){2-3} \cmidrule(lr){4-5} \cmidrule(lr){6-7}
 & 
  Micro &
  Macro &
  Micro &
  Macro &
  Micro &
  Macro  \\ \midrule
  \texttt{CEP}$_\text{independent}$ (LLaMA-3-8B) &
  99.4 &
  98.2 &
  69.7 &
  55.8 &
  85.8 &
  77.0 \\
\texttt{CEP}$_\text{integral}$ (LLaMA-3-8B) &
  97.3 &
  92.9 &
  68.4 &
  54.7 &
  85.1 &
  76.0 \\ \midrule
\multicolumn{7}{c}{\cellcolor{highlight_gray}\textbf{Planning}} \\ \midrule
 &
  \multirow{2}{*}{\begin{tabular}[c]{@{}c@{}}Delivery \\ Rate\end{tabular}} &
  \multicolumn{2}{c}{\begin{tabular}[c]{@{}c@{}}Commonsense \\ Pass Rate\end{tabular}} &
  \multicolumn{2}{c}{\begin{tabular}[c]{@{}c@{}}Hard Constraint \\ Pass Rate\end{tabular}} &
  \multirow{2}{*}{\begin{tabular}[c]{@{}c@{}}Final \\ Pass Rate\end{tabular}} \\ \cmidrule(lr){3-4} \cmidrule(lr){5-6}
 &
   &
  Micro &
  Macro &
  Micro &
  Macro &
   \\ \midrule
\texttt{CEP}$_\text{integral}$ &
  98.8 &
  64.3 &
  1.0 &
  19.2 &
  5.0 &
  0.1 \\
\texttt{CEP}$_\text{integral}$ w/o Clarification &
  93.3 &
  53.3 &
  0.3 &
  8.4 &
  3.1 &
  0 \\
\bottomrule
\end{tabular}
    \end{adjustbox}}
    \caption{Evaluation of the integral framework.}
    \label{tab:ablation_study}
    \vspace{-0.3cm}
\end{table}

%\subsection{Ablation Study} 
\subsubsection{Evaluation on the Integral Framework}
\label{sec:ablation_study}
%In the ablation study
Finally, we evaluate the integral \texttt{CEP} framework holistically rather than assessing the performance of each subtask independently. To this end, we adopt the execution interactions generated by the SOTA model in Table \ref{tab:tool_overall}, \texttt{CEP} (LLaMA-3-8B), for the subsequent clarification and the final planning. 
%Since real-time user simulation is not involved in our work, we still use the human-annotated ground truth for user-agent conversation. 
Results are reported in Table \ref{tab:ablation_study}.

% In the clarification subtask, we use \texttt{CEP} (LLaMA-3-8B) to complete the clarification need prediction along with question generation based on conversation ground truth and generated interaction chains, we find that the performance is slightly dropped because of incorrectness in the interaction, which points out the robustness of our problem setting. 

As for the clarification subtask, we use \texttt{CEP} (LLaMA-3-8B) based on the environment interaction generated by the execution agent and the current conversation history. Compared to the version supervised by ground-truth data, \textit{i.e.}, \texttt{CEP}$_\text{independent}$ (LLaMA-3-8B), the performance of \texttt{CEP}$_\text{integral}$ (LLaMA-3-8B) slightly drops due to incorrectly-predicted tool usage, highlighting the robustness of our proposed \texttt{CEP} framework. 

% Next, we focus on the planning subtask, investigating the impact of clarification. We select the ground truth conversation and tool interactions generated by \texttt{CEP}(LLaMA-3-8B). GPT-3.5-Turbo, identified as the best performer, was chosen as our planning agent. Observations indicate that without clarification, the planning agent fails to produce valid plans. Furthermore, clarification is essential for effective plan delivery and meeting constraints.
%Next, we focus on the planning subtask by investigating the impact of intention disambiguation. 
For the final planning subtask, we analyze the effect of user intention disambiguation by ablating the clarification agent. 
We adopt the best planning model from Table \ref{tab:plan_overall}, \textit{i.e.}, Direct (GPT-3.5), as the planning agent.  
%We feed the ground truth conversation and execution interactions generated by \texttt{CEP} (LLaMA-3-8B) into the best planning model, GPT-3.5. 
Specifically, we feed the initial instruction without clarification into the planning agent, denoted as w/o Clarification, for comparison.
The observations indicate that the planning agent fails to formulate valid plans without clarification, stating that user intention understanding is essential for effective plan delivery and meeting constraints.
To intuitively illustrate the importance of clarification in agent planning, we %demonstrate the constraint pass rate analysis in Section \ref{sec:detailed_analysis}, and 
present a detailed case study in Appendix \ref{app:case}. %[Move "present a detailed case study in Appendix \ref{app:case}" to the detailed analysis?]

\begin{figure}[t]
\setlength{\abovecaptionskip}{5pt}   
\setlength{\belowcaptionskip}{0pt}
\centering 
\includegraphics[width=0.5\textwidth]{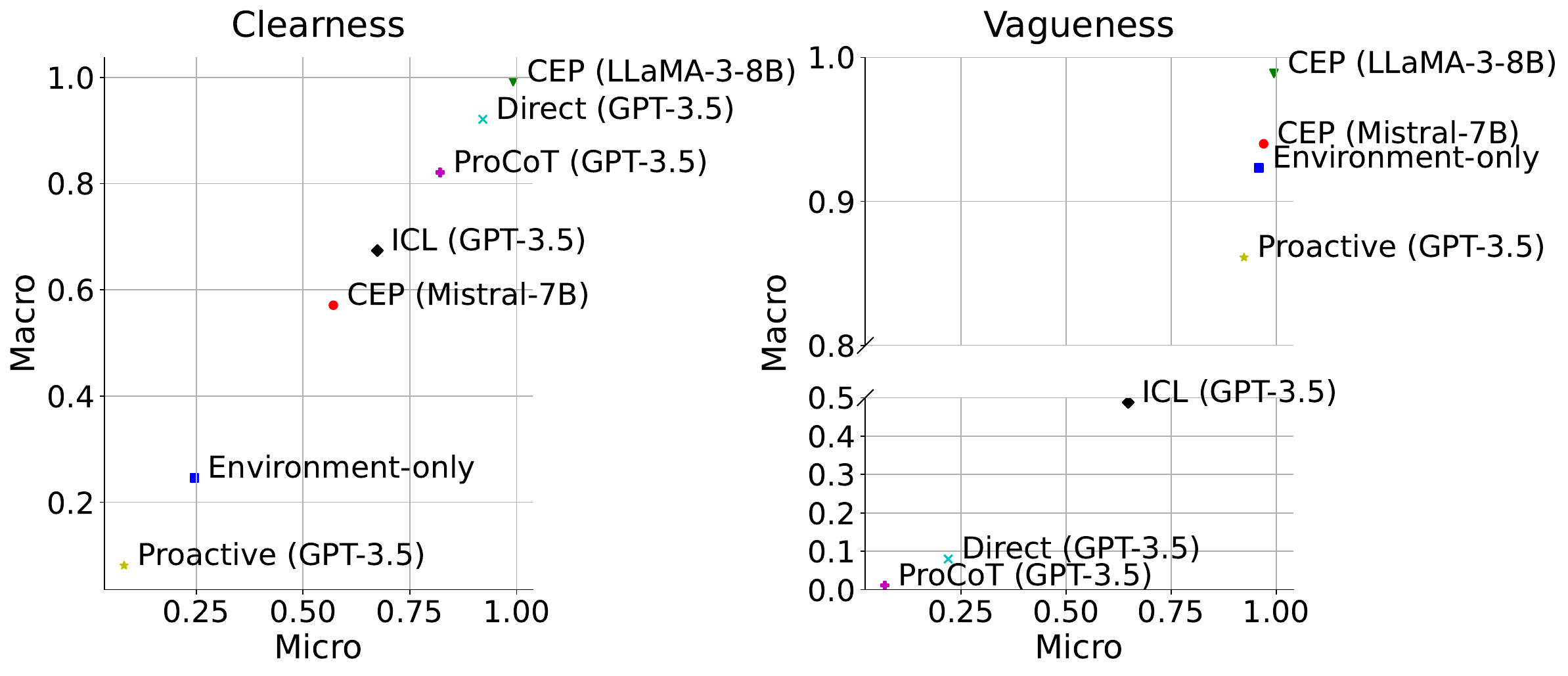}
\caption{Analysis of clarification recall.}
\label{fig:clarification_recall}
\vspace{-0.3cm}
\end{figure}

\subsection{Detailed Analysis}\label{sec:detailed_analysis}

\paragraph{Analysis of Clarification Recall}
We further examine the clarification accuracy from two angles: \textit{clearness} and \textit{vagueness} judgment recall (definitions are presented in Appendix \ref{app:metric}). It can be observed from Figure \ref{fig:clarification_recall} that: (1) \texttt{CEP} (LLaMA-3-8B) outperforms all other methods in both clearness and vagueness judgment. (2) Proactive (conversation-only) and environment-only approaches are inclined to request clarification, indicating that these agents struggle to make accurate judgments when either type of observation is absent. (3) ProCoT, Direct, and ICL (GPT-3.5) tend to not clarify the conversation, suggesting that full observations or invoking agents to engage in deeper reasoning might bring about hallucinations in judgment. (4) \texttt{CEP} (Mistral-7B) performs worse than \texttt{CEP} (LLaMA-3-8B) in both clearness and vagueness judgment, correlating with findings from static tool learning that LLaMA-3-8B's larger parameter size and better pretraining data lead to superior performance. %[Note that micro clearness recall is proportional to macro clearness recall because only the last conversation turn does not require clarification.]

\begin{table}[t]
    \setlength{\abovecaptionskip}{5pt}   
    \setlength{\belowcaptionskip}{0pt}
    \centering
    \setlength{\tabcolsep}{1mm}{
    \begin{adjustbox}{max width=0.48\textwidth}
    \begin{tabular}{lccccccccc}
\toprule
\multirow{2}{*}{Constraint Type} &
  \multicolumn{3}{c}{Greedy Search} &
  \multicolumn{3}{c}{\texttt{CEP}$_\text{integral}$} &
  \multicolumn{3}{c}{\texttt{CEP}$_\text{integral}$ w/o Clarif.} \\ \cmidrule(lr){2-4} \cmidrule(lr){5-7} \cmidrule(lr){8-10}
 &
 \multicolumn{1}{c}{Easy} &
  \multicolumn{1}{c}{Medium} &
  \multicolumn{1}{c}{Hard} &
  \multicolumn{1}{c}{Easy} &
  \multicolumn{1}{c}{Medium} &
  \multicolumn{1}{c}{Hard} &
  \multicolumn{1}{c}{Easy} &
  \multicolumn{1}{c}{Medium} &
  \multicolumn{1}{c}{Hard} \\ \midrule
\multicolumn{10}{c}{\cellcolor{highlight_gray}\textbf{Commonsense Constraint}} \\ \midrule
Within Sandbox             & 100 & 100 & 100 & 38.8 & 38.1 & 42.3 & 23.6 & 22.2 & 21.0 \\
Complete Information       & 100 & 100 & 100 & 89.7 & 89.8 & 74.6 & 67.0 & 55.0 & 48.0 \\
Within Current City        & 100 & 100 & 100 & 69.8 & 76.3 & 77.7 & 62.9 & 64.3 & 62.1 \\
Reasonable City Route      & 100 & 100 & 100 & 68.7 & 74.2 & 67.4 & 31.6 & 25.8 & 28.5 \\
Diverse Restaurants        & 0 & 0 & 0 & 65.2 & 69.4 & 72.4 & 66.7 & 72.1 & 73.7 \\
Diverse Attractions        & 100 & 100 & 100 & 93.1 & 93.7 & 92.2 & 89.7 & 90.4 & 89.0 \\
Non-conf. Transportation   & 93.4 & 92.5 & 91.8 & 74.4 & 70.3 & 89.3 & 64.9 & 55.6 & 68.0 \\
Minimum Nights Stay        & 20.4 & 24.9 & 22.9 & 6.0 & 5.1 & 6.0 & 31.6 & 30.0 & 34.5 \\ \midrule
\multicolumn{10}{c}{\cellcolor{highlight_gray}\textbf{Hard Constraint}}        \\ \midrule
Budget                     & 99.7 & 99.4 & 100 & 5.2 & 7.8 & 2.5 & 3.4 & 4.8 & 0.9   \\
Accommodation              & - & 41.3 & 32.0 & - & 33.8 & 34.8 & - & 16.4 & 14.4   \\
Cuisine                    & - & 5.9 & 0 & - & 32.8 & 19.8 & - & 8.4 & 8.4   \\
Transportation             & - & - & 55.0 & - & - & 37.2 & - & - & 15.6   \\ \bottomrule
\end{tabular}
    \end{adjustbox}}
    \caption{Analysis of plan constraints.}
    \label{tab:plan_constraints}
    \vspace{-0.3cm}
\end{table}

\paragraph{Analysis of Planning Constraints} 
% We present the detailed statistics of each commonsense and hard constraints for the planning subtask in Table \ref{tab:plan_constraints}. The descriptions of these constraint type are the same as TravelPlanner \cite{travelplanner}. We compare the greedy search, Mistral-7B, and GPT-3.5, and find (1) As the difficulty of the planning subtask increase, agents are more likely to fail to pass commonsense and hard constraints. (2) Agents perform better in within current city when the difficulty is raised. \dy{[so?]} (3) GPT-3.5 generally surpasses the Mistral-7B except the minimum nights stay. \dy{[so?]} (4) Mistral-7B demonstrates less capability in managing Non-conflict Transportation than GPT-3.5. \dy{[so?]}
We report the detailed planning constraint pass rates for Greedy Search, \texttt{CEP}$_\text{integral}$ with and without clarification in Table \ref{tab:plan_constraints}. The description of these constraints is shown in Appendix \ref{app:metric}. Our findings include: (1) Compared to Greedy Search, \texttt{CEP}$_\text{integral}$ underperforms on most constraints except for \textit{diverse restaurants} and \textit{cuisine}, highlighting the limitations of off-the-shelf language agents in Proactive Agent Planning. (2) Clarification plays a critical role in enabling agents to meet most commonsense and all hard constraints, as evidenced by the lower pass rates of \texttt{CEP}$_\text{integral}$ without clarification. 
However, an exception is the constraint \textit{minimum nights stay}, which checks whether the accommodation options entailed in a plan satisfy the minimum night requirements. 
This constraint demands mathematical understanding of the plan, while the lengthy context of clarification dialogues may introduce some noise into the planning process. 
\paragraph{Analysis of Framework Efficiency}
To delve into the time complexity and communication overhead of the proposed multi-agent framework, we measure the execution time for \texttt{CEP}$_\text{integral}$ from Table \ref{tab:ablation_study}. For each data sample involving multiple conversation turns and interaction iterations, we calculate the total time spent on three subtasks separately: an average of 14.17 seconds for tool learning, 5.40 seconds for clarification, and 8.78 seconds for planning. The experimental results demonstrate that \texttt{CEP}$_\text{integral}$ is efficient in real-world applications, as clarification requires significantly less time than tool learning and planning.

%\zx{However, two constraints are exceptions, including \textit{diverse restaurants} and \textit{minimum nights stay}. The constraint \textit{diverse restaurants} assesses the diversity of provided restaurant information within a plan, and the constraint \textit{minimum nights stay} checks whether the accommodation options entailed in a plan meet the minimum night requirements. In fact, neither constraint evaluates whether the plan fulfills the user's demands, which is the central objective of our proposed task. On the contrary, this discrepancy does not affect our conclusion and implies the limited planning ability of current LLM agents.}

\section{Conclusion}
%To explore the capabilities of current LLM-powered agents in managing clarification, execution, and planning subtasks over multiple turns, we introduce the Ask-Before-Plan dataset. This dataset focuses on predicting the need for clarification, generating questions based on user needs, interacting with the environment through instruction-following, and multi-step planning within constraints. To overcome challenges such as redundant reasoning, unnecessary inference time, and the inability of current agents to request clarification and execute functions in lengthy conversation contexts, we propose a novel framework called CEP (Clarification-Execution-Planning). This framework features trajectory tuning and \zx{the memory recollection mechanism}. We rigorously evaluate our Ask-Before-Plan dataset against extensive baselines across three independent subtasks. Our experimental results highlight the effectiveness of our proposed approach.

In this work, we introduce a new and practical problem of Proactive Agent Planning to address the challenges faced by LLM-powered language agents in handling unclear user instructions. To tackle these obstacles, we propose a novel multi-agent framework, \texttt{CEP}, which consists of clarification, execution, and planning agents. This framework features trajectory tuning and memory recollection to address challenges faced by current agents in requesting clarifications and executing tools within lengthy conversation contexts and complex environmental interactions. Additionally, we construct the first dataset specifically for studying the concerned problem, named Ask-before-Plan. Extensive evaluations and comprehensive analyses across diverse settings substantiate the effectiveness of the proposed \texttt{CEP} framework. %This work lays the foundation for future research in enhancing the proactivity and reliability of language agents.

\section*{Limitations}
\paragraph{User Simulation}
To simplify the evaluation on the clarification subtask, we employ static user simulation rather than real-time user simulation for dialogues. This approach allows us to focus exclusively on evaluating the agent's ability for clarification question generation, without the latent interference from incorrect user simulation. We believe that our proposed task could be easily extended to include dynamic real-time user simulation in future work. 
%\zx{While recent studies \cite{reliable-user-simulator} investigate real-time user simulation for proactive dialogues, our work does not involve this due to the complexity of implementation and challenges in ensuring its correctness. To simplify the clarification subtask, we employ static user simulation. This approach allows us to focus exclusively on evaluating the agent's ability for clarification question generation, without the latent interference from incorrect user simulation. We believe that our proposed task could be easily extended to include dynamic user simulation in future work.}

\paragraph{Agent Planning}
The effectiveness of our proposed framework, \texttt{CEP}, has not been verified on other benchmarks, as we are the first to introduce the problem of Proactive Agent Planning and there are no other suitable datasets available. Due to the limited computing resources, we are unable to conduct experiments with larger-size models. In addition, while cooperation and division of labor among different LLM-based agents improve planning capabilities in uncertain situations \cite{adaptive-team-building, multi-phase-planning}, our work does not thoroughly explore this. Our primary goal is to establish and benchmark this new and practical problem, as well as leave room for future research on this challenging problem.

\section*{Ethical Consideration}
The Ask-before-Plan dataset was built from the TravelPlanner dataset \cite{travelplanner}, which is publicly available. We have obtained permission from the authors of TravelPlanner to utilize the dataset for further construction.

\section*{Acknowledgement}
This research was supported by the Singapore Ministry of Education (MOE) Academic Research Fund (AcRF) Tier 1 grant (No. MSS24C004). This research/project is supported by A*STAR, CISCO Systems (USA) Pte. Ltd, and National University of Singapore under its Cisco-NUS Accelerated Digital Economy Corporate Laboratory (Award I21001E0002) and the National Research Foundation, Singapore, under its Industry Alignment Fund – Pre-positioning (IAF-PP) Funding Initiative. Any opinions, findings, conclusions, or recommendations expressed in this material are those of the author(s) and do not reflect the views of National Research Foundation, Singapore.

\bibliography{custom}

\begin{thebibliography}{44}
\providecommand{\natexlab}[1]{#1}

\bibitem[{Aliannejadi et~al.(2021)Aliannejadi, Kiseleva, Chuklin, Dalton, and Burtsev}]{emnlp21-opendomain-clarification}
Mohammad Aliannejadi, Julia Kiseleva, Aleksandr Chuklin, Jeff Dalton, and Mikhail Burtsev. 2021.
\newblock \href {https://doi.org/10.18653/v1/2021.emnlp-main.367} {Building and evaluating open-domain dialogue corpora with clarifying questions}.
\newblock In \emph{Proceedings of the 2021 Conference on Empirical Methods in Natural Language Processing}, pages 4473--4484. Association for Computational Linguistics.

\bibitem[{Andukuri et~al.(2024)Andukuri, Fr{\"a}nken, Gerstenberg, and Goodman}]{star-gate}
Chinmaya Andukuri, Jan-Philipp Fr{\"a}nken, Tobias Gerstenberg, and Noah~D Goodman. 2024.
\newblock \href {https://arxiv.org/abs/2403.19154} {Star-gate: Teaching language models to ask clarifying questions}.
\newblock \emph{ArXiv preprint}.

\bibitem[{Brown et~al.(2020)Brown, Mann, Ryder, Subbiah, Kaplan, Dhariwal, Neelakantan, Shyam, Sastry, Askell, Agarwal, Herbert{-}Voss, Krueger, Henighan, Child, Ramesh, Ziegler, Wu, Winter, Hesse, Chen, Sigler, Litwin, Gray, Chess, Clark, Berner, McCandlish, Radford, Sutskever, and Amodei}]{gpt3}
Tom~B. Brown, Benjamin Mann, Nick Ryder, Melanie Subbiah, Jared Kaplan, Prafulla Dhariwal, Arvind Neelakantan, Pranav Shyam, Girish Sastry, Amanda Askell, Sandhini Agarwal, Ariel Herbert{-}Voss, Gretchen Krueger, Tom Henighan, Rewon Child, Aditya Ramesh, Daniel~M. Ziegler, Jeffrey Wu, Clemens Winter, Christopher Hesse, Mark Chen, Eric Sigler, Mateusz Litwin, Scott Gray, Benjamin Chess, Jack Clark, Christopher Berner, Sam McCandlish, Alec Radford, Ilya Sutskever, and Dario Amodei. 2020.
\newblock \href {https://proceedings.neurips.cc/paper/2020/hash/1457c0d6bfcb4967418bfb8ac142f64a-Abstract.html} {Language models are few-shot learners}.
\newblock In \emph{Advances in Neural Information Processing Systems 33: Annual Conference on Neural Information Processing Systems 2020, NeurIPS 2020, December 6-12, 2020, virtual}.

\bibitem[{Chen et~al.(2024)Chen, Huang, Deng, Lei, Jin, Liu, and Chua}]{acl24-style}
Yue Chen, Chen Huang, Yang Deng, Wenqiang Lei, Dingnan Jin, Jia Liu, and Tat{-}Seng Chua. 2024.
\newblock \href {https://aclanthology.org/2024.findings-acl.632} {{STYLE:} improving domain transferability of asking clarification questions in large language model powered conversational agents}.
\newblock In \emph{Findings of the Association for Computational Linguistics, {ACL} 2024}, pages 10633--10649.

\bibitem[{Deng et~al.(2023{\natexlab{a}})Deng, Gu, Zheng, Chen, Stevens, Wang, Sun, and Su}]{mind2web}
Xiang Deng, Yu~Gu, Boyuan Zheng, Shijie Chen, Sam Stevens, Boshi Wang, Huan Sun, and Yu~Su. 2023{\natexlab{a}}.
\newblock Mind2web: Towards a generalist agent for the web.
\newblock \emph{Advances in Neural Information Processing Systems}.

\bibitem[{Deng et~al.(2023{\natexlab{b}})Deng, Lei, Lam, and Chua}]{proactive-dialogue-system}
Yang Deng, Wenqiang Lei, Wai Lam, and Tat-Seng Chua. 2023{\natexlab{b}}.
\newblock \href {https://doi.org/10.24963/ijcai.2023/738} {A survey on proactive dialogue systems: problems, methods, and prospects}.
\newblock In \emph{Proceedings of the Thirty-Second International Joint Conference on Artificial Intelligence}, IJCAI '23.

\bibitem[{Deng et~al.(2022)Deng, Lei, Zhang, Lam, and Chua}]{emnlp22-pacific}
Yang Deng, Wenqiang Lei, Wenxuan Zhang, Wai Lam, and Tat-Seng Chua. 2022.
\newblock \href {https://aclanthology.org/2022.emnlp-main.469} {{PACIFIC}: Towards proactive conversational question answering over tabular and textual data in finance}.
\newblock In \emph{Proceedings of the 2022 Conference on Empirical Methods in Natural Language Processing}, pages 6970--6984. Association for Computational Linguistics.

\bibitem[{Deng et~al.(2023{\natexlab{c}})Deng, Liao, Chen, Wang, Lei, and Chua}]{procot}
Yang Deng, Lizi Liao, Liang Chen, Hongru Wang, Wenqiang Lei, and Tat-Seng Chua. 2023{\natexlab{c}}.
\newblock \href {https://doi.org/10.18653/v1/2023.findings-emnlp.711} {Prompting and evaluating large language models for proactive dialogues: Clarification, target-guided, and non-collaboration}.
\newblock In \emph{Findings of the Association for Computational Linguistics: EMNLP 2023}. Association for Computational Linguistics.

\bibitem[{Deng et~al.(2024{\natexlab{a}})Deng, Zhang, Lin, Chen, Wen, and Chua}]{www24-agent-tutorial}
Yang Deng, An~Zhang, Yankai Lin, Xu~Chen, Ji{-}Rong Wen, and Tat{-}Seng Chua. 2024{\natexlab{a}}.
\newblock \href {https://doi.org/10.1145/3589335.3641240} {Large language model powered agents in the web}.
\newblock In \emph{Companion Proceedings of the {ACM} on Web Conference 2024, {WWW} 2024}, pages 1242--1245.

\bibitem[{Deng et~al.(2024{\natexlab{b}})Deng, Zhang, Zhang, Yuan, Ng, and Chua}]{self-map}
Yang Deng, Xuan Zhang, Wenxuan Zhang, Yifei Yuan, See-Kiong Ng, and Tat-Seng Chua. 2024{\natexlab{b}}.
\newblock \href {https://doi.org/10.18653/v1/2024.acl-long.477} {On the multi-turn instruction following for conversational web agents}.
\newblock In \emph{Proceedings of the 62nd Annual Meeting of the Association for Computational Linguistics (Volume 1: Long Papers)}, pages 8795--8812. Association for Computational Linguistics.

\bibitem[{Descartes and Cress(1998)}]{descartes}
R.~Descartes and D.A. Cress. 1998.
\newblock \emph{Discourse on Method (Third Edition)}.
\newblock HPC Classics Series. Hackett Publishing Company.

\bibitem[{Dreyfus(1992)}]{dreyfus}
Hubert~L. Dreyfus. 1992.
\newblock \emph{What Computers Still Can't Do: A Critique of Artificial Reason}.
\newblock MIT Press.

\bibitem[{Hu et~al.(2022)Hu, Shen, Wallis, Allen{-}Zhu, Li, Wang, Wang, and Chen}]{lora}
Edward~J. Hu, Yelong Shen, Phillip Wallis, Zeyuan Allen{-}Zhu, Yuanzhi Li, Shean Wang, Lu~Wang, and Weizhu Chen. 2022.
\newblock \href {https://openreview.net/forum?id=nZeVKeeFYf9} {Lora: Low-rank adaptation of large language models}.
\newblock In \emph{The Tenth International Conference on Learning Representations, {ICLR} 2022, Virtual Event, April 25-29, 2022}. OpenReview.net.

\bibitem[{Kojima et~al.(2023)Kojima, Gu, Reid, Matsuo, and Iwasawa}]{zero-shot-cot}
Takeshi Kojima, Shixiang~Shane Gu, Machel Reid, Yutaka Matsuo, and Yusuke Iwasawa. 2023.
\newblock Large language models are zero-shot reasoners.
\newblock \emph{Advances in neural information processing systems}.

\bibitem[{Kuhn et~al.(2022)Kuhn, Gal, and Farquhar}]{clam}
Lorenz Kuhn, Yarin Gal, and Sebastian Farquhar. 2022.
\newblock \href {https://arxiv.org/abs/2212.07769} {Clam: Selective clarification for ambiguous questions with generative language models}.
\newblock \emph{ArXiv preprint}.

\bibitem[{Li et~al.(2024{\natexlab{a}})Li, Wang, Zhang, Li, Lai, Kang, Ma, and Liu}]{hospital-agent}
Junkai Li, Siyu Wang, Meng Zhang, Weitao Li, Yunghwei Lai, Xinhui Kang, Weizhi Ma, and Yang Liu. 2024{\natexlab{a}}.
\newblock \href {https://arxiv.org/abs/2405.02957} {Agent hospital: A simulacrum of hospital with evolvable medical agents}.
\newblock \emph{ArXiv preprint}.

\bibitem[{Li et~al.(2024{\natexlab{b}})Li, Hua, Wang, Zhu, and Zhang}]{formal-llm}
Zelong Li, Wenyue Hua, Hao Wang, He~Zhu, and Yongfeng Zhang. 2024{\natexlab{b}}.
\newblock \href {https://arxiv.org/abs/2402.00798} {Formal-llm: Integrating formal language and natural language for controllable llm-based agents}.
\newblock \emph{ArXiv preprint}.

\bibitem[{Liao et~al.(2023)Liao, Yang, and Shah}]{sigar23-proactive-conversational-agent}
Lizi Liao, Grace~Hui Yang, and Chirag Shah. 2023.
\newblock \href {https://doi.org/10.1145/3539618.3594250} {Proactive conversational agents in the post-chatgpt world}.
\newblock In \emph{Proceedings of the 46th International ACM SIGIR Conference on Research and Development in Information Retrieval}, SIGIR '23. Association for Computing Machinery.

\bibitem[{Micikevicius et~al.(2022)Micikevicius, Stosic, Burgess, Cornea, Dubey, Grisenthwaite, Ha, Heinecke, Judd, Kamalu et~al.}]{fp8}
Paulius Micikevicius, Dusan Stosic, Neil Burgess, Marius Cornea, Pradeep Dubey, Richard Grisenthwaite, Sangwon Ha, Alexander Heinecke, Patrick Judd, John Kamalu, et~al. 2022.
\newblock \href {https://arxiv.org/abs/2209.05433} {Fp8 formats for deep learning}.
\newblock \emph{ArXiv preprint}.

\bibitem[{Papineni et~al.(2002)Papineni, Roukos, Ward, and Zhu}]{bleu}
Kishore Papineni, Salim Roukos, Todd Ward, and Wei-Jing Zhu. 2002.
\newblock \href {https://doi.org/10.3115/1073083.1073135} {{B}leu: a method for automatic evaluation of machine translation}.
\newblock In \emph{Proceedings of the 40th Annual Meeting of the Association for Computational Linguistics}, pages 311--318. Association for Computational Linguistics.

\bibitem[{Pyatkin et~al.(2023)Pyatkin, Hwang, Srikumar, Lu, Jiang, Choi, and Bhagavatula}]{clarify-delphi}
Valentina Pyatkin, Jena~D. Hwang, Vivek Srikumar, Ximing Lu, Liwei Jiang, Yejin Choi, and Chandra Bhagavatula. 2023.
\newblock \href {https://doi.org/10.18653/v1/2023.acl-long.630} {{C}larify{D}elphi: Reinforced clarification questions with defeasibility rewards for social and moral situations}.
\newblock In \emph{Proceedings of the 61st Annual Meeting of the Association for Computational Linguistics (Volume 1: Long Papers)}. Association for Computational Linguistics.

\bibitem[{Qian et~al.(2024)Qian, He, Zhuang, Deng, Qin, Cong, Zhang, Zhou, Lin, Liu, and Sun}]{tell-me-more}
Cheng Qian, Bingxiang He, Zhong Zhuang, Jia Deng, Yujia Qin, Xin Cong, Zhong Zhang, Jie Zhou, Yankai Lin, Zhiyuan Liu, and Maosong Sun. 2024.
\newblock \href {https://doi.org/10.18653/v1/2024.acl-long.61} {Tell me more! towards implicit user intention understanding of language model driven agents}.
\newblock In \emph{Proceedings of the 62nd Annual Meeting of the Association for Computational Linguistics (Volume 1: Long Papers)}, pages 1088--1113. Association for Computational Linguistics.

\bibitem[{Qin et~al.(2024)Qin, Liang, Ye, Zhu, Yan, Lu, Lin, Cong, Tang, Qian et~al.}]{toolllm}
Yujia Qin, Shihao Liang, Yining Ye, Kunlun Zhu, Lan Yan, Yaxi Lu, Yankai Lin, Xin Cong, Xiangru Tang, Bill Qian, et~al. 2024.
\newblock Toolllm: Facilitating large language models to master 16000+ real-world apis.
\newblock In \emph{ICLR 2024}.

\bibitem[{Rao and Daum{\'e}~III(2018)}]{acl18-ranking}
Sudha Rao and Hal Daum{\'e}~III. 2018.
\newblock \href {https://doi.org/10.18653/v1/P18-1255} {Learning to ask good questions: Ranking clarification questions using neural expected value of perfect information}.
\newblock In \emph{Proceedings of the 56th Annual Meeting of the Association for Computational Linguistics (Volume 1: Long Papers)}, pages 2737--2746. Association for Computational Linguistics.

\bibitem[{Schick et~al.(2023)Schick, Dwivedi-Yu, Dess{\`\i}, Raileanu, Lomeli, Hambro, Zettlemoyer, Cancedda, and Scialom}]{toolformer}
Timo Schick, Jane Dwivedi-Yu, Roberto Dess{\`\i}, Roberta Raileanu, Maria Lomeli, Eric Hambro, Luke Zettlemoyer, Nicola Cancedda, and Thomas Scialom. 2023.
\newblock Toolformer: Language models can teach themselves to use tools.
\newblock \emph{Advances in Neural Information Processing Systems}.

\bibitem[{Schmidgall et~al.(2024)Schmidgall, Ziaei, Harris, Reis, Jopling, and Moor}]{agent-clinic}
Samuel Schmidgall, Rojin Ziaei, Carl Harris, Eduardo Reis, Jeffrey Jopling, and Michael Moor. 2024.
\newblock \href {https://arxiv.org/abs/2405.07960} {Agentclinic: a multimodal agent benchmark to evaluate ai in simulated clinical environments}.
\newblock \emph{ArXiv preprint}.

\bibitem[{Shinn et~al.(2023)Shinn, Labash, and Gopinath}]{reflexion}
Noah Shinn, Beck Labash, and Ashwin Gopinath. 2023.
\newblock \href {https://arxiv.org/abs/2303.11366} {Reflexion: an autonomous agent with dynamic memory and self-reflection}.
\newblock \emph{ArXiv preprint}.

\bibitem[{Song et~al.(2024)Song, Liu, Zhang, Zhang, Luo, Wang, Wu, and Wang}]{adaptive-team-building}
Linxin Song, Jiale Liu, Jieyu Zhang, Shaokun Zhang, Ao~Luo, Shijian Wang, Qingyun Wu, and Chi Wang. 2024.
\newblock \href {https://arxiv.org/abs/2405.19425} {Adaptive in-conversation team building for language model agents}.
\newblock \emph{ArXiv preprint}.

\bibitem[{Wang et~al.(2024)Wang, Fang, Eisner, Van~Durme, and Su}]{tool-ste}
Boshi Wang, Hao Fang, Jason Eisner, Benjamin Van~Durme, and Yu~Su. 2024.
\newblock \href {https://arxiv.org/abs/2403.04746} {Llms in the imaginarium: tool learning through simulated trial and error}.
\newblock \emph{ArXiv preprint}.

\bibitem[{Wang et~al.(2023{\natexlab{a}})Wang, Xie, Jiang, Mandlekar, Xiao, Zhu, Fan, and Anandkumar}]{voyager}
Guanzhi Wang, Yuqi Xie, Yunfan Jiang, Ajay Mandlekar, Chaowei Xiao, Yuke Zhu, Linxi Fan, and Anima Anandkumar. 2023{\natexlab{a}}.
\newblock \href {https://arxiv.org/abs/2305.16291} {Voyager: An open-ended embodied agent with large language models}.
\newblock \emph{ArXiv preprint}.

\bibitem[{Wang et~al.(2023{\natexlab{b}})Wang, Ma, Feng, Zhang, Yang, Zhang, Chen, Tang, Chen, Lin, Zhao, Wei, and Wen}]{ruc-agent-survey}
Lei Wang, Chen Ma, Xueyang Feng, Zeyu Zhang, Hao Yang, Jingsen Zhang, Zhiyuan Chen, Jiakai Tang, Xu~Chen, Yankai Lin, Wayne~Xin Zhao, Zhewei Wei, and Ji{-}Rong Wen. 2023{\natexlab{b}}.
\newblock \href {https://arxiv.org/abs/2308.11432} {A survey on large language model based autonomous agents}.
\newblock \emph{ArXiv preprint}.

\bibitem[{Wen et~al.(2023)Wen, Fu, Li, Cai, Ma, Cai, Dou, Shi, He, and Qiao}]{dilu}
Licheng Wen, Daocheng Fu, Xin Li, Xinyu Cai, Tao Ma, Pinlong Cai, Min Dou, Botian Shi, Liang He, and Yu~Qiao. 2023.
\newblock \href {https://arxiv.org/abs/2309.16292} {Dilu: A knowledge-driven approach to autonomous driving with large language models}.
\newblock \emph{ArXiv preprint}.

\bibitem[{Wu et~al.(2023)Wu, Parish, Cheng, Min, Ammanabrolu, Ostendorf, and Hajishirzi}]{tacl2023-inscit}
Zeqiu Wu, Ryu Parish, Hao Cheng, Sewon Min, Prithviraj Ammanabrolu, Mari Ostendorf, and Hannaneh Hajishirzi. 2023.
\newblock \href {https://doi.org/10.1162/tacl_a_00559} {{I}n{SCI}t: Information-seeking conversations with mixed-initiative interactions}.
\newblock \emph{Transactions of the Association for Computational Linguistics}.

\bibitem[{Xi et~al.(2023)Xi, Chen, Guo, He, Ding, Hong, Zhang, Wang, Jin, Zhou et~al.}]{llm-agent-survey}
Zhiheng Xi, Wenxiang Chen, Xin Guo, Wei He, Yiwen Ding, Boyang Hong, Ming Zhang, Junzhe Wang, Senjie Jin, Enyu Zhou, et~al. 2023.
\newblock \href {https://arxiv.org/abs/2309.07864} {The rise and potential of large language model based agents: A survey}.
\newblock \emph{ArXiv preprint}.

\bibitem[{Xie and Zou(2024)}]{multi-phase-planning}
Chengxing Xie and Difan Zou. 2024.
\newblock \href {https://arxiv.org/abs/2405.18208} {A human-like reasoning framework for multi-phases planning task with large language models}.
\newblock \emph{ArXiv preprint}.

\bibitem[{Xie et~al.(2024)Xie, Zhang, Chen, Zhu, Lou, Tian, Xiao, and Su}]{travelplanner}
Jian Xie, Kai Zhang, Jiangjie Chen, Tinghui Zhu, Renze Lou, Yuandong Tian, Yanghua Xiao, and Yu~Su. 2024.
\newblock Travelplanner: A benchmark for real-world planning with language agents.
\newblock In \emph{ICML 2024}.

\bibitem[{Xu et~al.(2019)Xu, Wang, Tang, Duan, Yang, Zeng, Zhou, and Sun}]{emnlp19-kbqa}
Jingjing Xu, Yuechen Wang, Duyu Tang, Nan Duan, Pengcheng Yang, Qi~Zeng, Ming Zhou, and Xu~Sun. 2019.
\newblock \href {https://doi.org/10.18653/v1/D19-1172} {Asking clarification questions in knowledge-based question answering}.
\newblock In \emph{Proceedings of the 2019 Conference on Empirical Methods in Natural Language Processing and the 9th International Joint Conference on Natural Language Processing (EMNLP-IJCNLP)}, pages 1618--1629. Association for Computational Linguistics.

\bibitem[{Yao et~al.(2022)Yao, Chen, Yang, and Narasimhan}]{webshop}
Shunyu Yao, Howard Chen, John Yang, and Karthik Narasimhan. 2022.
\newblock Webshop: Towards scalable real-world web interaction with grounded language agents.
\newblock \emph{Advances in Neural Information Processing Systems}.

\bibitem[{Yao et~al.(2023)Yao, Zhao, Yu, Du, Shafran, Narasimhan, and Cao}]{react}
Shunyu Yao, Jeffrey Zhao, Dian Yu, Nan Du, Izhak Shafran, Karthik~R Narasimhan, and Yuan Cao. 2023.
\newblock React: Synergizing reasoning and acting in language models.
\newblock In \emph{The Eleventh International Conference on Learning Representations}.

\bibitem[{Zamani et~al.(2020)Zamani, Dumais, Craswell, Bennett, and Lueck}]{www20-ir-clarification}
Hamed Zamani, Susan~T. Dumais, Nick Craswell, Paul~N. Bennett, and Gord Lueck. 2020.
\newblock \href {https://doi.org/10.1145/3366423.3380126} {Generating clarifying questions for information retrieval}.
\newblock In \emph{{WWW} '20: The Web Conference 2020, Taipei, Taiwan, April 20-24, 2020}, pages 418--428. {ACM} / {IW3C2}.

\bibitem[{Zhang et~al.(2024{\natexlab{a}})Zhang, Deng, Lin, Chen, Wen, and Chua}]{sigir24-agent-tutorial}
An~Zhang, Yang Deng, Yankai Lin, Xu~Chen, Ji{-}Rong Wen, and Tat{-}Seng Chua. 2024{\natexlab{a}}.
\newblock \href {https://doi.org/10.1145/3626772.3661375} {Large language model powered agents for information retrieval}.
\newblock In \emph{Proceedings of the 47th International {ACM} {SIGIR} Conference on Research and Development in Information Retrieval, {SIGIR} 2024}, pages 2989--2992.

\bibitem[{Zhang et~al.(2024{\natexlab{b}})Zhang, Qin, Deng, Huang, Lei, Liu, Jin, Liang, and Chua}]{acl24-clamber}
Tong Zhang, Peixin Qin, Yang Deng, Chen Huang, Wenqiang Lei, Junhong Liu, Dingnan Jin, Hongru Liang, and Tat-Seng Chua. 2024{\natexlab{b}}.
\newblock Clamber: A benchmark of identifying and clarifying ambiguous information needs in large language models.
\newblock In \emph{ACL 2024}.

\bibitem[{Zhong et~al.(2024)Zhong, Guo, Gao, Ye, and Wang}]{memory-bank}
Wanjun Zhong, Lianghong Guo, Qiqi Gao, He~Ye, and Yanlin Wang. 2024.
\newblock Memorybank: Enhancing large language models with long-term memory.
\newblock In \emph{Proceedings of the AAAI Conference on Artificial Intelligence}, 17.

\bibitem[{Zhu et~al.(2023)Zhu, Chen, Tian, Tao, Su, Yang, Huang, Li, Lu, Wang et~al.}]{minecraft-ghost}
Xizhou Zhu, Yuntao Chen, Hao Tian, Chenxin Tao, Weijie Su, Chenyu Yang, Gao Huang, Bin Li, Lewei Lu, Xiaogang Wang, et~al. 2023.
\newblock \href {https://arxiv.org/abs/2305.17144} {Ghost in the minecraft: Generally capable agents for open-world enviroments via large language models with text-based knowledge and memory}.
\newblock \emph{ArXiv preprint}.

\end{thebibliography}

\appendix
\section{Dataset Specifications}\label{app:dataset}
To adapt TravelPlanner to our proposed task, we modify the dataset and environment for proactive information seeking. In the following sections, we share our implementation of the \textbf{Ask-before-Plan} dataset along with the environment.

\subsection{Indefinite Detail Formation}\label{app:indefinite_detail}

\begin{figure}[htbp]
    \centering
    \includegraphics[width=0.48\textwidth]{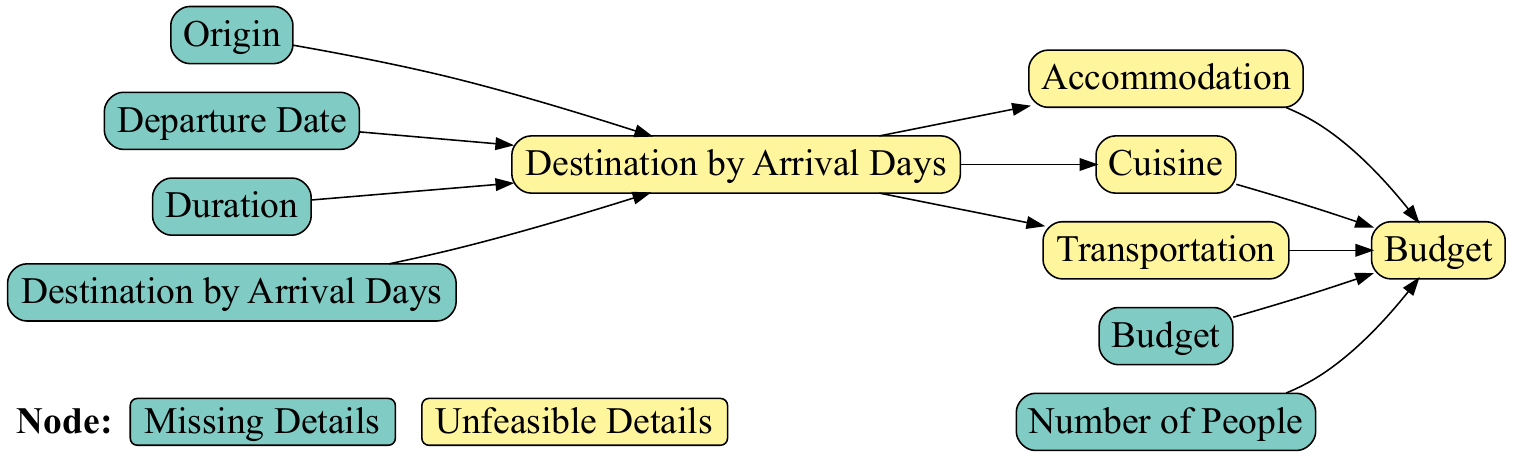}
    \caption{Dependency graph for indefinite details.}
    \label{fig:dependency_graph}
\end{figure}

To simplify the dataset and tailor it to our task, we specify the destination cities and related arrival days, rather than providing the destination state and the number of cities to visit. In this case, we present only the departure date for each data sample in lieu of the date range. Besides, we combine the room type and the house rule under local constraints as the \textit{accommodation}. Table~\ref{table:dataset_element} shows the description of elements in the dataset. To construct the unfeasible details, we apply the following procedures for each feature:
\begin{itemize}[leftmargin=*]
    \item Accommodation: We search all combinations of room types and house rules for every destination in the environment to identify candidates with no matching information, and then randomly select one of these accommodation preferences as the unfeasible detail.
    \item Cuisine: We iterate all possible cuisine preferences in each destination. If some cuisine preferences are not available in any destination, we randomly choose one of them as the unfeasible detail.
    \item Transportation: We assign the preference "No flight" or "No self-driving" as the unfeasible detail if the corresponding transportation mode is not available for the route described in the instruction. It is guaranteed in the environment that both preferences will not be unavailable simultaneously.
    \item Destination by Arrival Days: We randomly modify one of the destinations to be unreachable from the previous city in the travel route. To preserve the consistency of destinations, cities within the same state as the original destination are preferentially chosen for modification.
    \item Budget: We begin by estimating the lower bound of the budget for the user's travel plan by summing the accommodation and transportation expenses. To determine the minimum accommodation cost, we select the least expensive accommodation in the given city that satisfies the \textit{minimum nights} constraint. The minimum transportation cost is calculated by choosing the cheapest option between "No flight" and "No self-driving". Ultimately, we subtract a random integer, ranging from 10 to 10\% of the final estimated lower bound, from the total approximate budget.
\end{itemize}
Upon obtaining all the indefinite details, we select between 0 and 3 for each instruction. Afterwards, we apply the topological sort algorithm, based on the directed acyclic graph illustrated in Figure \ref{fig:dependency_graph}, to regulate the dependency between these details. Eventually, we utilize the prompts outlined in \ref{app:data_annotation_prompts} and employ \texttt{gpt-3.5-turbo-0125} to rewrite the original instruction.

\begin{table*}[t]
\small
\centering
\begin{tabularx}{\textwidth}{X}
\toprule
\textbf{Destination by Arrival Days} \\[1ex]
Definition: \textit{An array depicts the destination city with the day to arrive.}  \\
Example in the Natural Language Instruction: \textit{We plan to visit Dallas on the 1st day and Houston on the 3rd day...} \\[1ex]
\midrule
\textbf{Duration} \\[1ex]
Definition: \textit{The number of travel days.}  \\
Example in the Natural Language Instruction: \textit{We are planning a 3-day trip...}\\[1ex]
\midrule
\textbf{Departure Date} \\[1ex]
Definition: \textit{Date of departure from the origin city.}  \\
Example in the Natural Language Instruction: \textit{Would you be able to organize a trip on March 18th, 2022...}\\[1ex]
\midrule
\textbf{Number of People} \\[1ex]
Definition: \textit{The total number of individuals on the trip.}  \\
Example in the Natural Language Instruction: \textit{Please organize a trip for 2 individuals...}\\[1ex]
\midrule
\textbf{Budget} \\[1ex]
Definition: \textit{The budget for the trip in integers.}  \\
Example in the Natural Language Instruction: \textit{... we have a budget of \$1600...}\\[1ex]
\midrule
\textbf{Origin} \\[1ex]
Definition: \textit{The departure city of the trip.}  \\
Example in the Natural Language Instruction: \textit{Would you mind creating a travel plan starting from St. Louis...}\\[1ex]
\midrule
\textbf{Accommodation} \\[1ex]
Definition: \textit{An optional array describes the accommodation preferences, consisting of the room type and the house rule. Possible values for the room type include "shared room", "not shared room", "private room", and "entire room". Possible values for the house rule include "parties", "smoking", "children under 10", "pets", and "visitors".}  \\
Example in the Natural Language Instruction: \textit{... Our accommodation preference is a place that allows parties...}\\[1ex]
\midrule
\textbf{Transportation} \\[1ex]
Definition: \textit{An optional string indicates transportation preferences, such as "No flight" and "No self-driving".}  \\
Example in the Natural Language Instruction: \textit{... Additionally, we prefer not to use flights for transportation...}\\[1ex]
\midrule
\textbf{Cuisine} \\[1ex]
Definition: \textit{An optional array represents cuisine preferences, with possible values including "Chinese", "American", "Italian", "Mexican", "Indian", "Mediterranean", and "French".}  \\
Example in the Natural Language Instruction: \textit{... We are interested in exploring Chinese, Indian, Mexican, and Italian cuisines...}\\[1ex]
\bottomrule
\end{tabularx}
\caption{Description of elements in the dataset.}
\label{table:dataset_element}
\end{table*}
% \zx{Adaptation of the dataset and environment: to make the environment of TravelPlanner fit our task, we delete the details in TravelPlanner like the visiting city number. In our task, we specify the city user plan to go, not given a destination state for the agent to plan. Besides, we combine the room types and house rules for simplicity. To deal with the information like dates, we divide the dates attribute into two parts: departure date, and arrival days for each city, and we combine the destination names with its arrival days. To make the environment consistent with our modified task in this new task, we remove the tools like Notebook Write, and City Search. Besides, we add the new tool named Budget Estimator, and also revise the required parameters to fit our setting.} \dy{[this paragraph can be removed. But use a section in the appendix to introduce each type of the missing details / unfeasible details, and provide an example for each.]}

\subsection{Proactive Conversation Generation}\label{app:proactive_conversation}
We apply \texttt{gpt-3.5-turbo-0125} to implement conversation generation, with the specific prompts utilized detailed in Section \ref{app:data_annotation_prompts}. Notably, to accurately guide GPT-3.5 in generating role-playing dialogues, we incorporate both XML and Python-type structures into the prompts. In the quality control phase, we initially use keyword matching to ensure that relevant keywords are included in both the questions and responses. Finally, we select 20\% of the data points at random for proofreading.

\subsection{Environment Settings}\label{app:environment}
To align the TravelPlanner environment with our proposed task, we modify it by removing tools such as \textit{NotebookWrite} and \textit{CitySearch}. We introduce a new tool named \textit{BudgetEstimator} to accommodate the unfeasible detail, \textit{budget}. Furthermore, we adjust the required parameters of each tool to better suit our dataset setup. The description of modified tools in our environment is as follows:

\begin{itemize}[leftmargin=*]
    \item \textbf{AccommodationSearch}: Instead of merely searching for accommodation options in the specified city, an optional parameter, \textit{filters}, is introduced to describe the user's accommodation preferences. This parameter corresponds directly to the dataset element \textit{accommodation}.
    \item \textbf{RestaurantSearch}: We add an optional parameter, \textit{cuisines}, to capture the user's culinary preferences, corresponding to the dataset element \textit{cuisine}. This enhancement enables the tool to list all restaurants in the selected city and verify the availability of the user's preferred cuisine.
    \item \textbf{BudgetEstimator}: This tool is used to calculate the minimum budget required for the trip based on the cheapest accommodation and transportation options. The parameters include the departure city, destination cities, dates of arrival in each city, and the number of people.
\end{itemize}

\section{Details of Experimental Setups}\label{app:experiment}

\subsection{Baselines}\label{app:baseline}
We describe the details of the adopted baselines. The prompts for the clarification, execution, and planning subtasks are exhibited in Sections \ref{app:clarification_prompts}, \ref{app:execution_prompts}, and \ref{app:planning_prompts}.

\paragraph{Clarification} For the clarification subtask, we adopt the following baselines:
\begin{itemize}[leftmargin=*]
    \item Environment-only: To implement the generation of clarification questions based on environmental feedback, we prepare predefined questions tailored to two scenarios: one where a tool invocation is not recorded in the interaction log, and another where a tool invocation suggests that the current travel plan conversation is infeasible. For each interaction history, if one of these scenarios occurs, we add the corresponding questions to the candidate pool and finally sample one question from this pool to ask the user. 
    \item Proactive \cite{procot}: This method provides alternative choices for agents to decide whether to ask a question for disambiguation or take no action. 

    \item ProCoT \cite{procot}: This approach distinguishes itself from the \textit{Proactive} baseline by involving reasoning about the next action prior to making a decision.
    \item Direct: Given the conversation history and the current interaction log, the agent is prompted to finish Clarification Need Prediction and Clarification Question Generation without additional strategies. In the conversation-only setting, environmental information is excluded from the prompts.
    \item ICL \cite{gpt3}: Based on Direct prompting, two examples for clarification need prediction and one for question generation are selected from the training dataset and incorporated into the context.
\end{itemize}

In Clarification Need Prediction, we record the first "Yes" or "No" from the agent's response as the final judgment. If the response does not contain "Yes" or "No", we consider the prediction to be "No". 

\paragraph{Execution} We adopt the following baselines in the execution subtask:

\begin{itemize}[leftmargin=*]
    \item Brute-force: We provide the brute-force solution, as a representative of the traditional algorithm, for the execution subtask. As the name suggests, this solution invokes all tools, regardless of syntax correctness, based on the information from the conversation history.
    \item Direct: In this method, the agent is prompted to generate all possible tool invocations according to the conversation history. For GPT-3.5, we provide an example to enable in-context learning.
    \item ToolLLM \cite{toolllm}: Following the Direct prompting, we train ToolLLaMA on our Ask-before-Plan dataset in the static setting for 3 epochs, after pretraining on ToolBench for 2 epochs, to match the total number of training epochs with \texttt{CEP}.
    \item ReAct \cite{react}: This prompting strategy necessitates that the agent integrate environmental observations into its reasoning process. In the execution subtask, the ReAct agent produces step-by-step tool commands based on the conversation history and its previous interactions with the environment, cycling through \textit{thought}, \textit{action}, and \textit{observation}.
    \item Reflexion \cite{reflexion}: This approach is similar to ReAct, but the agent is required to reflect on the incorrect tool invocations and provide a high-level explanation of the error in order to improve its reasoning.
\end{itemize}

\paragraph{Planning} We use the same baseline setting as described in the TravelPlanner sole-planning mode, which is outlined as follows:

\begin{itemize}[leftmargin=*]
    \item Greedy Search: We borrow the greedy search algorithm from TravelPlanner and adapt it to our planning subtask. To illustrate, we remove destination selection since our dataset specifies the city to visit for 5- or 7-day travel plans.
    \item Direct: In this baseline, the agent directly generates the plan given the conversation and interaction history.
    \item CoT \cite{zero-shot-cot}: Compared with the direct baseline, we add the prompt "Let's think step by step" to elicit reasoning capability.
    \item ReAct \cite{react}: Similar to ReAct in the execution subtask, we only provide the agent with a cost calculation environment to assist in generating the travel plan.
    \item Reflexion \cite{reflexion}: Based on ReAct, we introduce the self-reflection module, similar to the execution subtask, for guiding the agent to generate a valid travel plan.
\end{itemize}

Unlike TravelPlanner, which first generates a natural language plan and then uses GPT-4 to convert it into a JSON format, our planning subtask requires directly generating a JSON travel plan.

\subsection{Evaluation Metrics}\label{app:metric}
Considering the conversational nature of clarification and execution subtasks, we introduce two different evaluation strategies: \textit{micro} and \textit{macro}. The micro strategy calculates the average per conversation turn, whereas the macro strategy calculates the average per data sample. In the planning subtask, the definition of \textit{micro} and \textit{macro} strategies is consistent with TravelPlanner \cite{travelplanner}, which is the ratio of passed constraints to the total number of constraints and the ratio of passed constraints among all tested plans, respectively.

\paragraph{Clarification}
We incorporate the following metrics to evaluate the clarification subtask:

\begin{itemize}[leftmargin=*]
    \item Clarification Accuracy: The percentage of correct Clarification Need Prediction. A prediction is considered correct if the agent accurately identifies when a conversation requires clarification.
    \item Clearness Recall: The ratio of conversation turns predicted to be clear to the total number of turns that do not require clarification. This measures the agent's ability to identify unambiguous user inquiries.
    \item Vagueness Recall: The ratio of conversation turns predicted to be vague to the total number of turns that require clarification. This metric evaluates the agent's capability to identify ambiguous user inquiries.
    \item Rule-based Score: The percentage of generated questions containing the predefined keywords for respective indefinite details.
    \item BLEU \cite{bleu}: This metric measures the similarity between the generated question and the ground truth in terms of n-gram overlap. We set the maximum n-gram order to 4 without smoothing.
    \item GPT Score: We utilize the OpenAI model \texttt{gpt-4-turbo-2024-04-09} with zero temperature to judge the correctness of the generated questions. Prompts are shown in Section \ref{app:clarification_prompts}.
\end{itemize}

Note that the last three metrics are exclusively used to assess the correctness of the generated questions. Consequently, we perform these calculations only when the Clarification Need Prediction is correct. Given that multiple indefinite details may require clarification within a single conversation turn, for these metrics, we calculate the highest score across all candidate indefinite details.

\paragraph{Execution} For conversation turn $t$, we denote the generated tool calls as $F_{\textrm{gen}}^t$ and the ground truth tool calls as $F_{\textrm{gt}}^t$. We define $\textrm{set}(\cdot)$ as a function that eliminates duplicate elements from an array. Following \citet{tool-ste}, the metrics for the execution subtask are as follows:

\begin{itemize}[leftmargin=*]
    \item Well-formed: The percentage of $F_{\text{gen}}^t$ without syntax errors.
    \item API Match: This metric calculates the ratio of matched API names between $\textrm{set}(F_{\text{gen}}^t)$ and $\textrm{set}(F_{\text{gt}}^t)$ to the total number in $\textrm{set}(F_{\text{gt}}^t)$.
    \item Repeat Rate: The percentage of repeated tool calls in $F_{\text{gen}}^t$.
    \item Correctness: We measure the precision, recall, and F1 score between $\textrm{set}(F_{\text{gen}}^t)$ and $\textrm{set}(F_{\text{gt}}^t)$.
    \item Pass Rate: This metric in the micro strategy evaluates the full match of tool execution, \textit{i.e.}, $\textrm{set}(F_{\text{gen}}^t) = \textrm{set}(F_{\text{gt}}^t)$. This macro metric evaluates that $\forall t, \textrm{set}(F_{\text{gen}}^t) = \textrm{set}(F_{\text{gt}}^t)$.
\end{itemize}

\paragraph{Planning}
In this subtask, we adapt the evaluation metrics from TravelPlanner \cite{travelplanner} to our setting, shown as follows:
\begin{itemize}[leftmargin=*]
    \item Delivery Rate: The percentage of valid JSON plans within a limited interaction step.
    \item Commonsense Constraint Pass Rate: This metric evaluates whether the planning agent could incorporate commonsense, incorporating \textit{within sandbox}, \textit{complete information}, \textit{within current city}, \textit{reasonable city route}, \textit{diverse restaurants}, \textit{diverse attractions}, \textit{non-conflict transportation}, \textit{minimum nights stay} into the plan without explicit instructions.
    \item Hard Constraint Pass Rate: This metric measures whether a generated plan satisfies all hard constraints, including \textit{budget}, \textit{accommodation}, \textit{cuisine}, \textit{transportation}.
    \item Final Pass Rate: The metric reports the percentage of valid JSON plans that satisfy all constraints.
\end{itemize}
To differentiate, we combine the hard constraints \textit{room rule} and \textit{room type} from TravelPlanner into \textit{accommodation} to fit our dataset. Since we specify the destinations and their arrival days in our dataset, we do not evaluate the number of visiting cities or the state of destinations in our commonsense constraint assessment. Additionally, the \textit{minimum nights stay} constraint in our setting is designed to verify whether the accommodation option provided by the agent aligns with the user's expected duration of stay in a city.
\subsection{Implementation Details}\label{app:implementation}
We choose Mistral-7B-Instruct\footnote{\url{https://huggingface.co/mistralai/Mistral-7B-Instruct-v0.2}}, ToolLLaMA-2-7B\footnote{\url{https://huggingface.co/ToolBench/ToolLLaMA-2-7b-v2}}, and LLaMA-3-8B-Instruct\footnote{\url{https://huggingface.co/meta-llama/Meta-Llama-3-8B-Instruct}} for finetuning and inference using a single NVIDIA A5000 24GB. To facilitate finetuning, we apply LoRA \cite{lora} in 4-bit quantization and train for 5 epoches with a maximum context length of 4096. To optimize the open-source LLM inference, FP8 E5M2 KV Cache \cite{fp8} is exploited. 

In dynamic execution and planning evaluations, we exclude LLaMA-3-8B due to the limited context window length. To maximize the utilization of the context window, we truncate the environment feedback in \textit{Clarification} and \textit{Execution} subtasks to 3 items. In terms of dynamic prompting strategies such as ReAct and Reflexion, we set the maximum interaction steps to 30 and the maximum action retries to 3. Our temperature is set to 0 for all experiments. 

For the integral framework, we feed the clarification need prediction and execution interactions generated by \texttt{CEP} (LLaMA-3-8B), along with the ground truth conversation, into the final planning.

\section{Additional Study on Dynamic Execution Agents}\label{app:dynamic_execution}
We discuss the time complexity and execution status of the dynamic execution agents in this section.

\subsection{Analysis of Time Complexity}\label{app:time_complexity}
To precisely outline the memory recollection, let $C_t$ denote the conversation at turn $t$, associated with the clarified detail set $D'_t = \{d_0, d_1, \ldots, d_t\}$, where $d_0$ pertains to the details entailed in the initial query $q_0$. It can be observed that $C_{t-1}$ and $C_t$ share the common clarified details, defined as $D'_{t} - D'_{t-1} = \{d_{0}, d_1, \ldots, d_{t-1}\}$. We postulate that an agent, operating under deterministic parameters, consistently replicates specific errors when invoking tools on these shared details. In the Reflexion framework, if the rationales generated during conversation turn $t$ are represented by $R_t$, then the cumulative number of generated rationales across $T$ turns is expressed as $\sum_{t=1}^T|R_t|$. Conversely, in memory recollection, as we \textit{reuse} the past rationales, this can be reduced to $\max_{t=1}^T |R_t| = |R_T|$. In conclusion, the memory recollection mechanism effectively reduces the number of generated rationales, thereby diminishing the time complexity of rationale inference from linear to constant time.

\begin{figure}[t]
\setlength{\abovecaptionskip}{5pt}   
\setlength{\belowcaptionskip}{0pt}
\centering 
\includegraphics[width=0.48\textwidth]{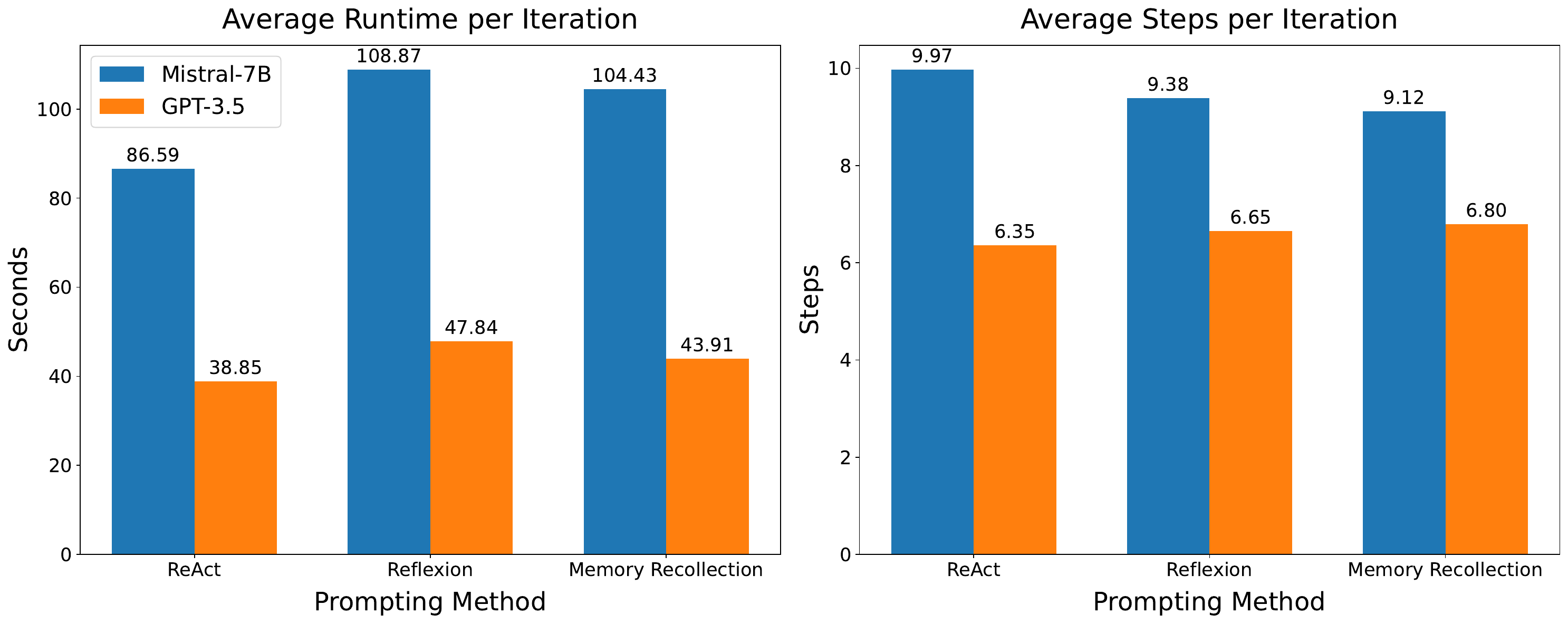}
\caption{Average runtime and steps in dynamic execution.}
\label{fig:time_complexity}
\vspace{-0.3cm}
\end{figure}

Figure \ref{fig:time_complexity} showcases a comparison in average runtime and execution steps between ReAct, Reflexion, and Memory Recollection. We observe that Reflexion consumes more time than ReAct for both Mistral-7B and GPT-3.5, while the Memory Recollection module alleviates this effect. Moreover, Mistral-7B, on average, takes longer inference time than GPT-3.5. We also examine the mean execution steps per iteration for Mistral-7B and GPT-3.5, revealing a starkly contrasting trend: Mistral-7B shows a decrease in steps from ReAct to Memory Recollection, while GPT-3.5 exhibits an increase. This phenomenon, detailed in Table \ref{tab:tool_overall}, suggests that more execution steps lead to improved performance, as the agent engages more with the environment and gathers more information.

\subsection{Analysis of Execution Status}\label{app:execution_status}

\begin{figure}[t]
\setlength{\abovecaptionskip}{5pt}   
\setlength{\belowcaptionskip}{0pt}
\centering 
\includegraphics[width=0.48\textwidth]{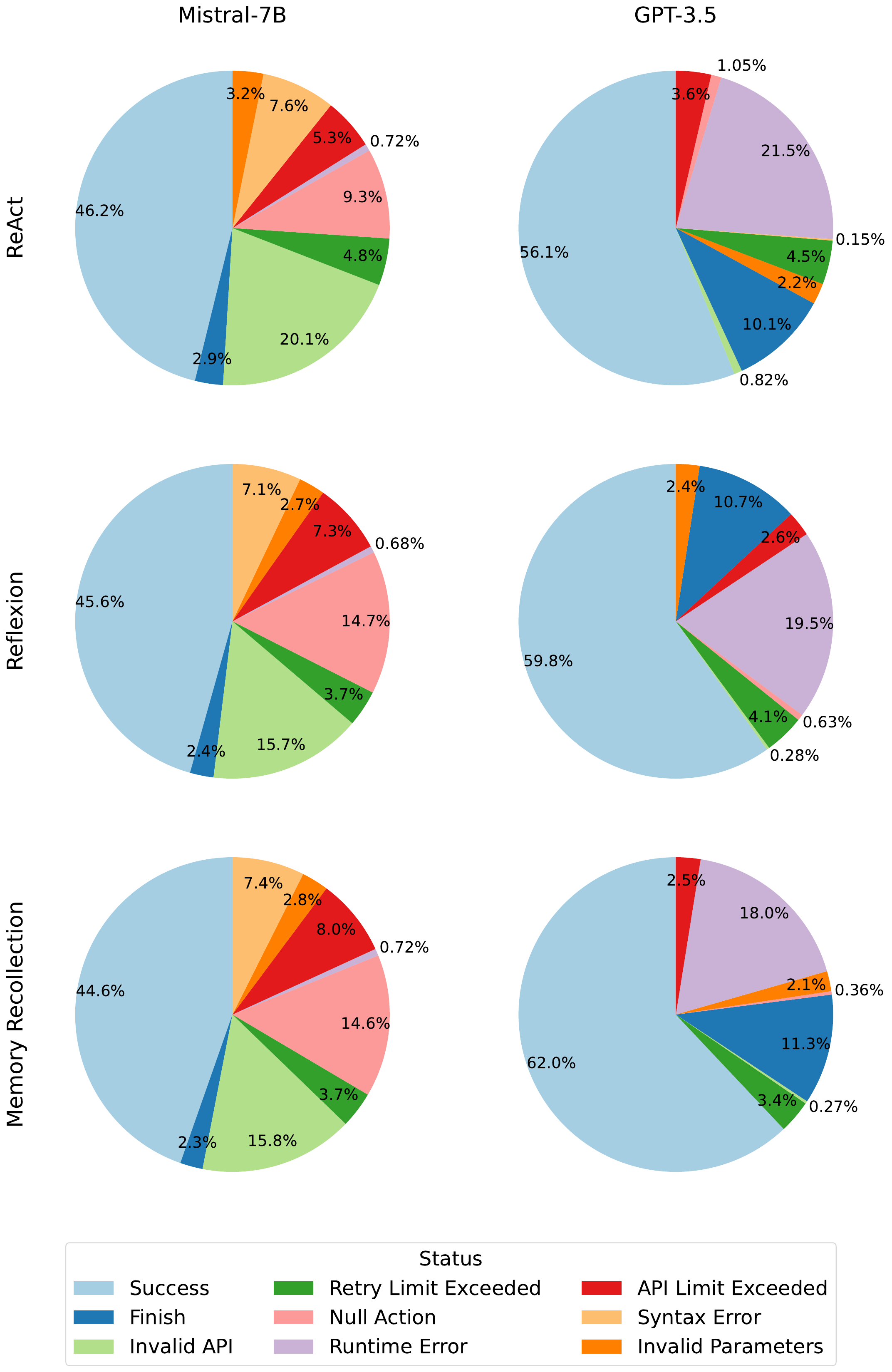}
\caption{Execution status distributions. Statuses with proportions less than 0.1\% are ignored.}
\label{fig:execution_status}
\vspace{-0.3cm}
\end{figure}

Figure \ref{fig:execution_status} presents the distribution of each execution status for Mistral-7B and GPT-3.5. From a holistic perspective, we notice that Reflexion eliminates the errors consisting of \textit{Invalid API}, \textit{Retry Limit Exceeded}, and \textit{Invalid Parameters}. In the case of Mistral-7B, prompting strategies like Reflexion and Memory Recollection lead to lower \textit{Success} and \textit{Finish} rates, while the probability of exceptions such as \textit{API Limit Exceeded} and \textit{Null Action} rises. This is in line with our observation in Table \ref{tab:tool_overall}. In contrast, the performance improvements of GPT-3.5 are reflected in higher \textit{Success} and \textit{Finish} rates, and lower error rates.

\lstset{
  breaklines=true,
  xleftmargin=3pt,
  breakindent=0pt,
  columns=fullflexible,
  mathescape,
  aboveskip=6pt,
  belowskip=12pt
}

\tcbset{
  boxrule=1.5pt,
  after=\vspace{0.75cm}
}

% Define and initialize the counter for the tcolorbox
\newcounter{snippetcounter}
\setcounter{snippetcounter}{1}  % Start counting from 0 and increment before use

% Define a new tcolorbox environment for snippets
\newtcolorbox[auto counter]{snippetbox}[3][]{%
    title=\textsc{Snippet~\thesnippetcounter: #2},
    before upper={\stepcounter{snippetcounter}},  % Increment counter right before the box content
    label={#3},  % Use the correct counter for labeling
    #1  % Allow passing additional options when using the box
}

\section{Case Study} \label{app:case}
We display two detailed case studies to explore the impact of clarification on subsequent planning subtasks. We compare plans generated by \texttt{CEP}$_\text{integral}$ with and without clarification in Tables \ref{tab:missing_detail_plan} and \ref{tab:unfeasible_detail_plan}. Note that each table includes the initial user instruction, the type of indefinite detail, the question generated by finetuned LLaMA-3-8B, the ground truth response, and the planning outcome. Discrepancies resulting from the clarification are highlighted in yellow. From Table \ref{tab:missing_detail_plan}, it is evident that without asking for the missing destination detail, the agent incorrectly confines the itinerary to the origin city, Nashville, on the first day, leading to a confusing itinerary. In Table \ref{tab:unfeasible_detail_plan}, if the agent overlooks that the previously specified accommodation preferences in Houston are unavailable and fails to inquire about the user's opinion, an incomplete travel plan will be generated.

\begin{table*}[ht]
\small
\centering
\begin{tabularx}{\textwidth}{lp{6cm}p{6cm}}
\toprule
\multicolumn{3}{X}{\textbf{Instruction}: Could you suggest a 5-day travel itinerary for a solo traveler starting from Nashville, departing on March 13th, 2022, with a budget of \$2123?} \\
\multicolumn{3}{l}{\textbf{Missing Detail}: Destination by arrival days.} \\
\multicolumn{3}{X}{\textbf{Question}: Which cities are you interested in visiting during your 5-day trip, and on which specific days do you plan to arrive in each city?} \\
\multicolumn{3}{X}{\textbf{Response}: I can't wait to explore the vibrant streets of San Diego on day 1 and then soak up the glitz and glamour of Los Angeles on day 3!}       \\

\midrule
             & \texttt{CEP}$_\text{integral}$                         & \texttt{CEP}$_\text{integral}$ w/o Clarification                        \\[1ex] \midrule
\textbf{Day}            & 1                           & 1                                      \\[0.5ex] \midrule
\textbf{Current City}   & from Nashville to San Diego & Nashville                              \\[0.5ex]\midrule
\rowcolor{highlight_yellow}\textbf{Transportation} & Flight Number: F3973475, from Nashville to San Diego, Departure Time: 09:14, Arrival Time: 11:22, Cost: \$485                            & N/A                                   \\[0.5ex] \midrule
\rowcolor{highlight_yellow}\textbf{Breakfast}      & Open Yard, San Diego                            & Biscuit Love Gulch, Nashville          \\[0.5ex]\midrule
\rowcolor{highlight_yellow}\textbf{Attraction}     & Cabrillo National Monument, San Diego; La Jolla Shores Park, San Diego                            & Country Music Hall of Fame and Museum, Nashville; Ryman Auditorium, Nashville  \\[0.5ex] \midrule
\rowcolor{highlight_yellow}\textbf{Lunch}          & Bikaner Sweets, San Diego                            & Hattie B's Hot Chicken, Nashville      \\[0.5ex] \midrule
\rowcolor{highlight_yellow}\textbf{Dinner}         & Chawlas 2, San Diego                            & Martin's Bar-B-Que Joint, Nashville    \\[0.5ex]\midrule
\rowcolor{highlight_yellow}\textbf{Accommodation}  & Spacious Room in Large 2 Bedroom Prewar Apartment, San Diego                            & Cozy Loft in East Nashville, Nashville \\[1.5ex]
\bottomrule
\end{tabularx}
\caption{Comparison of travel plans for the missing detail, destination by arrival days.}
\label{tab:missing_detail_plan}
\end{table*}

\begin{table*}[bp]
\small
\centering
\begin{tabularx}{\textwidth}{lp{6cm}p{6cm}}
\toprule
\multicolumn{3}{X}{\textbf{Instruction}: I am planning a solo trip starting from Bloomington. The itinerary includes visiting Dallas on day 1, Amarillo on day 3, and Houston on day 5. I will be departing on March 5th, 2022. I am looking for shared room accommodations that allow pets.} \\
\multicolumn{3}{l}{\textbf{Unfeasible Detail}: Accommodation.} \\
\multicolumn{3}{X}{\textbf{Question}: It seems like the accommodation preferences you've specified might not be feasible for this trip. Is there a different type of accommodation you would consider for your solo trip?} \\
\multicolumn{3}{X}{\textbf{Response}: Any accommodation preference is acceptable.}       \\
\midrule
              & \texttt{CEP}$_\text{integral}$                         & \texttt{CEP}$_\text{integral}$ w/o Clarification                        \\[1ex] \midrule
\textbf{Day}            & 5                           & 5                                      \\[0.5ex] \midrule
\textbf{Current City}   & from Amarillo to Houston & from Amarillo to Houston                              \\[0.5ex]\midrule
\textbf{Transportation} & Flight Number: F3822285, from Amarillo to Houston, Departure Time: 06:31, Arrival Time: 08:05, Cost: \$103                            & Taxi: from Amarillo to Houston, Duration: 8 hours 49 mins, Distance: 965 km, Cost: 965                                  \\[0.5ex] \midrule
\textbf{Breakfast}      & N/A                           & Jalapenos, Houston          \\[0.5ex]\midrule
\textbf{Attraction}     & Downtown Aquarium, Houston; Space Center Houston, Houston                            & Downtown Aquarium, Houston; Space Center Houston, Houston  \\[0.5ex] \midrule
\textbf{Lunch}          & Jalapenos, Houston                            & The BrewMaster - The Mix Fine Dine, Houston      \\[0.5ex]  \midrule
\textbf{Dinner}         & Royal Mart, Houston                            & "Pebble Street, Houston    \\[0.5ex]\midrule \rowcolor{highlight_yellow}
\textbf{Accommodation}  & Superhost 3 bedroom DISCOUNT, Houston                            & N/A \\[1.5ex]
\bottomrule
\end{tabularx}
\caption{Comparison of travel plans for the unfeasible detail, accommodation.}
\label{tab:unfeasible_detail_plan}
\end{table*}
\section{Prompt List} \label{app:prompts}
\subsection{Data Annotation}\label{app:data_annotation_prompts}
We present the prompts for original instruction revision during the indefinite detail formation stage in Table \ref{prompt:original_instruction}. Prompts and examples involved in proactive conversation generation are shown in Tables \ref{prompt:simulated_conversation}, \ref{prompt:missing_detail_example}, and \ref{prompt:unfeasible_detail_example}.

\subsection{Clarification Subtask}\label{app:clarification_prompts}
Since the clarification subtask necessitates the interaction history from the execution stage as input, we provide the relevant tool description in Snippet \ref{prompt:clarification_tool_description}. Prompts of conversation-only methods, including Proactive and ProCoT, as well as the direct strategy, are exhibited in Snippets \ref{prompt:conversation_only} and \ref{prompt:direct_clarification}. Furthermore, we also specify the prompts for GPT-4 evaluation in Table \ref{prompt:clarification_evaluation}.

\subsection{Execution Subtask}\label{app:execution_prompts}
The tool description used in execution is showcased in Snippet \ref{prompt:execution_tool_description}. We demonstrate the prompts for static and dynamic settings in Snippets \ref{prompt:static_execution} and \ref{prompt:dynamic_execution}. The error prompts for dynamic interaction are presented in Snippet \ref{prompt:dynamic_execution_error}.

\subsection{Planning Subtask}\label{app:planning_prompts}
The prompts for static planning, dynamic planning, and errors in dynamic interaction are displayed in Snippets \ref{prompt:static_planning}, \ref{prompt:dynamic_planning}, and \ref{prompt:dynamic_planning_error}.

{
\onecolumn
\begin{table}[t]
\small
\centering
\begin{tabularx}{\textwidth}{p{1.25cm}X}
    \toprule
    \multicolumn{1}{c}{\textbf{Role}} & \multicolumn{1}{c}{\textbf{Content}} \\
    \midrule
    system & Given a JSON with data where 'org' denotes the departure city, 'dest' contains the destination cities to be visited along with the arrival day, and 'housing' contains the room type and house rules. Please generate a natural language query beginning with '\{begin\_word\}' and including the arrival day if available. \\ \midrule
    user & 
    JSON:\newline\{"org": "Gulfport", "dest": ["Charlotte on day 1"], "departure\_date": "2022-03-05", "budget": 1800\}\newline QUERY: \\
    & \\
    assistant & Please design a travel plan from Gulfport heading to Charlotte on day 1, departing on March 5th, 2022, with a budget of \$1800. \\ \midrule
    user & JSON:\newline\{"org": "Omaha", "days": 5, "people\_number": 7, "housing": ["pets"], "budget": 35300\}\newline QUERY: \\
    & \\
    assistant & Could you provide a 5-day travel itinerary for a group of seven, starting in Omaha? Our budget is set at \$35,300, and it's essential that our accommodations be pet-friendly since we're bringing our pets. \\ \midrule
    user & JSON:\newline \{"dest": ["Atlanta on day 1", "Augusta on day 3", "Savannah on day 5"], "days": 7, "departure\_date": "2022-03-01", "people\_number": 2, "housing": ["entire room", "parties"], "cuisine": ["Italian", "Indian"], "transportation": "no self-driving"\}\newline QUERY: \\
    & \\
    assistant & I'm looking for a week-long trip for 2 individuals. We plan to visit Atlanta on the 1st day, Augusta on the 3rd day, and Savannah on the 5th day. For accommodations, we'd prefer an entire room that allows parties. We intend to depart on March 1st, 2022, and will navigate our journey without self-driving. In terms of food, we're enthusiasts of Italian food, and we'd also appreciate indulging in genuine Indian cuisine. \\

    \bottomrule
\end{tabularx}
\caption{Prompts for original instruction revision.}
\label{prompt:original_instruction}
\end{table}

\begin{table}[t]
\small
\centering
\begin{tabularx}{\textwidth}{lX}
    \toprule
    \multicolumn{1}{c}{\textbf{Type}} & \multicolumn{1}{c}{\textbf{System Message}} \\
    \midrule
    Missing Detail Observation &  You are an intelligent agent designed to interact with users to clarify and specify their requests. When given a user's initial query and a specific detail that is missing, your task is to generate a natural, conversational question to obtain that specific missing information from the user. Your response should use coreference or omission to refer back to the initial query, minimizing direct repetition of its details.
    \\ \midrule
    Unfeasible Detail Observation & You are an intelligent agent designed to interact with users to clarify and specify their requests based on the search results from external tools. When a user's initial query includes unfeasible details, as determined by these search results, your task is to inform the user that the initial query cannot be fulfilled due to these details. Then, generate a natural, conversational question to obtain an alternative option from the user. Your response should use coreference or omission to refer back to the initial query, minimizing direct repetition of its details. 
    \\ \midrule
    Succinct User Response & You are an intelligent agent designed to act as a real human user talking to a travel agent. When asked for details or clarifications about your travel plans, reply succinctly and directly using only the provided draft answers, ensuring your responses are natural, human-like, and creative without repeating the question.
    \\ \midrule
    Passionate User Response & You are an intelligent agent designed to act as a real human user talking to a travel agent. When asked for details or clarifications about your travel plans, reply diversely and passionately using only the provided draft answers, ensuring your responses are natural, human-like, and creative without repeating the question.
    \\ \bottomrule
\end{tabularx}
\caption{System messages for simulated conversations.}
\label{prompt:simulated_conversation}
\end{table}

\begin{table}[t]
\small
\centering
\begin{tabularx}{\textwidth}{X}
\toprule
\textbf{Destination by Arrival Days} \\[1ex]
Instruction: \textit{Could you provide a 5-day travel itinerary for a group of seven, starting in Omaha? Our budget is set at \$35,300, and it's essential that our accommodations be pet-friendly since we're bringing our pets.}  \\
Observation: \textit{\missingdetail{Destinations and arrive days of the trip}} \\
Question: \textit{Could you specify which cities you plan to visit during the trip, and the specific days you plan to arrive in each city?} \\
Thought: \textit{\draftanswer{destinations\_and\_arrive\_days = ...}}\\
Answer: \textit{We plan to go Seattle on the 1st day.}\\[1ex]
\midrule
\textbf{Duration} \\[1ex]
Instruction: \textit{Please design a travel plan departing from Gulfport and heading to Charlotte on day 1, departing on March 5th, 2022, with a budget of \$1800.}  \\
Observation: \textit{\missingdetail{Number of days for the trip}}\\
Question: \textit{Sorry for the confusion, but could you please clarify the number of days you plan to spend on this trip?} \\
Thought: \textit{\draftanswer{number\_of\_days\_for\_trip = ...}}\\
Answer: \textit{5 day.}\\[1ex]
\midrule
\textbf{Departure Date} \\[1ex]
Instruction: \textit{I'm looking for a week-long trip for 2 individuals. We plan to visit Atlanta on the 1st day, Augusta on the 3rd day, and Savannah on the 5th day. For accommodations, we'd prefer an entire room that allows parties. We don't like driving during our journey.}  \\
Observation: \text{\missingdetail{Departure date of the trip}} \\
Question: \textit{I think I missed the departure date for your trip. Could you provide that information?} \\
Thought: \textit{\draftanswer{departure\_date = ...}}\\
Answer: \textit{March 1st, 2022.} \\[1ex]
\midrule
\textbf{Number of People} \\[1ex]
Instruction: \textit{Could you provide a 5-day travel itinerary, starting in Omaha? Our budget is set at \$35,300, and it's essential that our accommodations be pet-friendly since we're bringing our pets.}  \\
Observation: \textit{\missingdetail{Number of people on the trip}} \\
Question: \textit{I'm not sure about the number of people in your group. Would you mind sharing that information?} \\
Thought: \textit{\draftanswer{total\_number\_of\_people\_including\_me = ...}}\\
Answer: \textit{We are a group of seven.}\\[1ex]
\midrule
\textbf{Budget} \\[1ex]
Instruction: \textit{Please design a travel plan departing from Gulfport and heading to Charlotte on day 1, departing on March 5th, 2022.}  \\
Observation: \textit{\missingdetail{Budget of the trip}} \\
Question: \textit{It seems you haven't mentioned the expected budget for this trip. Could you provide that information?} \\
Thought: \textit{\draftanswer{budget\_of\_trip = ...}}\\
Answer: \textit{Our budget for this trip is \$36,000.}\\[1ex]
\midrule
\textbf{Origin} \\[1ex]
Instruction: \textit{We plan to visit South Bend on the 1st day, Ithaca on the 3rd day departing on March 5th, 2022 for a 5-day trip. Our budget is \$1800.}  \\
Observation: \textit{\missingdetail{Departure city of the trip}} \\ 
Question: \textit{Sorry, I am not sure about the departure city for your trip. Could you provide that information?} \\
Thought: \textit{\draftanswer{departure\_city = ...}}\\
Answer: \textit{Ann Arbor.}\\[1ex]
\bottomrule
\end{tabularx}
\caption{Examples of missing details in proactive conversation generation.}
\label{prompt:missing_detail_example}
\end{table}

\begin{table}[t]
\small
\centering
\begin{tabularx}{\textwidth}{X}
\toprule
\textbf{Budget} \\[1ex]
Instruction: \textit{Could you provide a 5-day travel itinerary for a group of seven, starting in Omaha? Our budget is set at \$35,300, and it's essential that our accommodations be pet-friendly since we're bringing our pets.}  \\
Observation: \textit{\searchresult{The current budget ... is not sufficient for the trip. The minimum budget required is ...}} \\
Question: \textit{It seems like the budget you've mentioned might be insufficient for your trip. Would you be open to considering a higher budget?} \\
Thought: \textit{\draftanswer{Adjust the budget to ...}}\\
Answer: \textit{My Budget could be \$40,000.}
\\[1ex]

\midrule
\textbf{Accommodation} \\[1ex]
Instruction: \textit{Please design a travel plan departing from Gulfport and heading to Charlotte on day 1, departing on March 5th, 2022. I prefer to stay in private rooms that allow parties.  }\\
Observation: \textit{\searchresult{The current accommodation preference ... is not available for the trip.}} \\
Question: \textit{The accommodation preference you've mentioned might be a bit restrictive. Would you be open to considering other options?} \\
Thought: \textit{\draftanswer{...}}\\
Answer: \textit{Sure, how about any rooms that allow parties?}\\[1ex]

\midrule
\textbf{Transportation} \\[1ex]
Instruction: \textit{I'm looking for a week-long trip for 2 individuals. We plan to visit Atlanta on the 1st day, Augusta on the 3rd day, and Savannah on the 5th day. For accommodations, we'd prefer an entire room that allows parties. We don't like driving during our journey.}  \\
Observation: \textit{\searchresult{The current transportation preference ... is not available for the trip.}}\\
Question: \textit{It seems like navigating the entire trip without self-driving might be quite challenging. Would you be open to considering other transportation options?} \\
Thought: \textit{\draftanswer{...}}\\
Answer: \textit{Any transportation preference is acceptable.}\\[1ex]

\midrule
\textbf{Cuisine} \\[1ex]
Instruction: \textit{Could you provide a 5-day travel itinerary for a group of seven, starting in Omaha? We plan to go Seattle on the 1st day. We enjoy Chinese and Italian cuisine.}  \\
Observation: \textit{\searchresult{The current cuisine preference ... is not available in any of the cities you plan to visit.}} \\
Question: \textit{Sorry, I do not find any Chinese restaurants in cities you plan to visit. Could you provide alternative cuisine preferences?}\\
Thought: \textit{\draftanswer{...}}\\
Answer: \textit{I am open to any cuisine preference.}\\[1ex]

\midrule
\textbf{Destination by Arrival Days} \\[1ex]
Instruction: \textit{We plan to visit South Bend on the 1st day, Ithaca on the 3rd day departing from Richmond on March 5th, 2022 for a 5-day trip. Our budget is \$1800.}  \\
Observation: \textit{\searchresult{... is unreachable by all available transportation methods.}} \\
Question: \textit{Sorry, I couldn't find a way to arrive in Ithca. Could you provide an alternative destination?} \\
Thought: \textit{\draftanswer{My alternative destination is ...}}\\
Answer: \textit{Charlotte could be my alternative destination.}\\[1ex]
\bottomrule
\end{tabularx}
\caption{Examples of unfeasible details in proactive conversation generation.}
\label{prompt:unfeasible_detail_example}
\end{table}

\begin{table*}[b]
\small
\centering
\begin{tabularx}{\textwidth}{lX}
    \toprule
    \multicolumn{1}{c}{\textbf{Message Type}} & \multicolumn{1}{c}{\textbf{Content}} \\
    \midrule
    System Message &  You are a helpful assistant skilled at evaluating questions.
    \\ \midrule
    User Message for Missing Details & Please check if the following question exclusively asks for [$\textit{...}$], rather than [$\textit{...}$]. Provide a simple "Yes" or "No" answer.\newline Question: [$\textit{...}$]
    \\ \midrule
    User Message for Unfeasible Details & Please check if the question indicates that the initial [$\textit{...}$] is/are unfeasible and requests changes to the [$\textit{...}$], rather than [$\textit{...}$]. Provide a simple "Yes" or "No" answer.\newline Question: [$\textit{...}$]
    \\ \bottomrule
\end{tabularx}
\caption{Prompts for clarification evaluation.}
\label{prompt:clarification_evaluation}
\end{table*}
\twocolumn
}
{\small
\onecolumn

\begin{snippetbox}[colback=background_gray,colframe=frame_gray,rounded corners]{Tool Description in Clarification}{prompt:clarification_tool_description}
\begin{lstlisting}[breaklines=true, xleftmargin=0pt, breakindent=0pt, columns=fullflexible, mathescape]
1. AccommodationSearch(city, filters)
- Description: Discover accommodations in your desired city with specific filters.
- Parameters: 
  - city (str, required): The name of the city where you're seeking accommodation.
  - filters (list[str], required): A list of filters to refine your search. Choices include "shared room", "not shared room", "private room", "entire room", "parties", "smoking", "children under 10", "pets", "visitors". If the user does not specify any accommodation filters, assign an empty list "[]" to the parameter.

2. RestaurantSearch(city, cuisines)
- Description: List all restaurants in your chosen city, regardless of cuisine type, and check if any of the cuisines you specify are unavailable there.
- Parameters:
  - city (str, required): The name of the city where you're seeking restaurants.
  - cuisines (list[str], required): A list of desired cuisines to check for availability. Available options include "Chinese", "American", "Italian", "Mexican", "Indian", "Mediterranean", and "French". If the user does not specify any cuisines, assign an empty list "[]" to the parameter.

3. AttractionSearch(city)
- Description: Find attractions in a city of your choice.
- Parameters: 
  - city (str, required): The name of the city where you're seeking attractions.

4. DistanceMatrix(origin, destination, mode)
- Description: Estimate the distance, time, and cost between two cities.
- Parameters:
  - origin (str, required): The departure city of your journey.
  - destination (str, required): The destination city of your journey.
  - mode (str, required): The method of transportation. Choices include "self-driving" and "taxi".

5. FlightSearch(origin, destination, date):
- Description: A flight information retrieval tool.
- Parameters:
  - origin (str, required): The city you'll be flying out from.
  - destination (str, required): The city you aim to reach.
  - date (str, required): The date of your travel in "YYYY-MM-DD" format.

6. BudgetEstimator(origin, destination, dates, people_number)
- Description: Calculate the minimal estimated budget for the trip. Use this tool to verify if the budget provided by the user is sufficient.
- Parameters:
  - origin (str, required): The departure city of your trip.
  - destination (list[str], required): A list of cities you plan to visit during your trip.
  - dates (list[str], required): A list of dates corresponding to the departure from the origin and each of the destinations. The first date is the departure from the origin, and each subsequent date corresponds to the departure from the respective city in the destination list. The last date in this list is the departure from the final destination back to the origin city. All dates should be formatted as "YYYY-MM-DD".
  - people_number (int, required): The number of people on the trip.
\end{lstlisting}

\end{snippetbox}
\twocolumn
}

{\small
\onecolumn

\begin{snippetbox}[colback=background_red,colframe=frame_red,rounded corners]{Conversation-only Prompts}{prompt:conversation_only}
\textbf{System Message}
\begin{lstlisting}
Your current task is to determine the user's intentions and satisfy their needs based on the provided conversation between the user and the assistant.
\end{lstlisting}
\textbf{Proactive User Message}

\begin{lstlisting}
Conversation: [$\textit{...}$]
Based on the conversation, you have two options: ask a clarifying question or take no action. Choose the appropriate option to formulate your answer, starting with either "The clarifying question is" or "No action should be taken". Note that if the user does not specify details regarding accommodation, cuisine, or transportation in the initial query, it indicates that the user does not have specific preferences that need clarification in the following conversation.
\end{lstlisting}

\textbf{ProCoT User Message}

\begin{lstlisting}
Conversation: [$\textit{...}$]
Based on the conversation, first determine whether the user's request is ambiguous. A request is ambiguous if it contains missing or unfeasible details. If it is ambiguous, ask a clarifying question. If it is not ambiguous, no action is needed. Your response should start with an analysis of the ambiguity and then conclude with either "Therefore, the request is not ambiguous. No action should be taken." or "Therefore, the request is ambiguous. The clarifying question is". Note that if the user does not specify details regarding accommodation, cuisine, or transportation in the initial query, it indicates that the user does not have specific preferences that need clarification in the following conversation.
\end{lstlisting}
\end{snippetbox}

\begin{snippetbox}[colback=background_red,colframe=frame_red,rounded corners,after=\vspace{0cm}]{Direct Clarification Prompts}{prompt:direct_clarification}
\textbf{System Message}
\begin{lstlisting}
Your current task is to determine the user's intentions and satisfy their needs based on the provided conversation between the user and the assistant, along with the interaction trajectory involving tool use between the agent and the environment. The interaction trajectory includes the following tools:

[$\textit{tool description}$]

If certain tools are not called in the interaction trajectory, it indicates a lack of the required parameters needed to call those tools. For each category, including accommodations, dining, attractions, transportation, and budget, at least one relevant tool should be used during the interaction to gather sufficient information to help the user provide a clear and feasible request. Note that if the user does not specify details regarding accommodation, cuisine, or transportation in the initial query, it indicates that the user does not have specific preferences that need clarification in the following conversation.
\end{lstlisting}
\textbf{Clarification Need Prediction}
\begin{lstlisting}
Conversation: [$\textit{...}$]
Interaction trajectory: [$\textit{...}$]
Please determine whether the user's request needs clarification. A request needs clarification if the user's intention contains missing or unfeasible details based on the tool parameters and call results in the interaction trajectory. If the user's intention requires clarification, answer "Yes"; if it is clear and feasible, answer "No".
Answer:
\end{lstlisting}
\textbf{Clarification Question Generation}
\begin{lstlisting}
Please ask the user one clarification question to gather more information about a specific detail. Do not attempt to solve the task.
Question:
\end{lstlisting}

\end{snippetbox}
\twocolumn
}

{\small
\onecolumn
\begin{snippetbox}[colback=background_gray,colframe=frame_gray,rounded corners,after=\vspace{0cm}]{Tool Description in Execution}{prompt:execution_tool_description}
\begin{lstlisting}[breaklines=true, xleftmargin=0pt, breakindent=0pt, columns=fullflexible, mathescape]
1. AccommodationSearch(city, filters)
- Description: Discover accommodations in your desired city with specific filters.
- Parameters: 
  - city (str, required): The name of the city where you're seeking accommodation.
  - filters (list[str], required): A list of filters to refine your search. Choices include "shared room", "not shared room", "private room", "entire room", "parties", "smoking", "children under 10", "pets", "visitors". If the user does not specify any accommodation filters, assign an empty list "[]" to the parameter.
- Example: AccommodationSearch("Berlin", ["private room", "parties"]) would return private rooms in Berlin that allow parties.

2. RestaurantSearch(city, cuisines)
- Description: List all restaurants in your chosen city, regardless of cuisine type, and check if any of the cuisines you specify are unavailable there.
- Parameters:
  - city (str, required): The name of the city where you're seeking restaurants.
  - cuisines (list[str], required): A list of desired cuisines to check for availability. Available options include "Chinese", "American", "Italian", "Mexican", "Indian", "Mediterranean", and "French". If the user does not specify any cuisines, assign an empty list "[]" to the parameter.
- Example: RestaurantSearch("Dublin", ["Chinese", "Italian", "French"]) returns all restaurants in Dublin and informs you if any of the Chinese, Italian, or French cuisines are unavailable.

3. AttractionSearch(city)
- Description: Find attractions in a city of your choice.
- Parameters: 
  - city (str, required): The name of the city where you're seeking attractions.
- Example: AttractionSearch("London") would return attractions in London.

4. DistanceMatrix(origin, destination, mode)
- Description: Estimate the distance, time, and cost between two cities.
- Parameters:
  - origin (str, required): The departure city of your journey.
  - destination (str, required): The destination city of your journey.
  - mode (str, required): The method of transportation. Choices include "self-driving" and "taxi".
- Example: DistanceMatrix("Paris", "Lyon", "self-driving") would provide driving distance, time, and cost between Paris and Lyon.

5. FlightSearch(origin, destination, date):
- Description: A flight information retrieval tool.
- Parameters:
  - origin (str, required): The city you'll be flying out from.
  - destination (str, required): The city you aim to reach.
  - date (str, required): The date of your travel in "YYYY-MM-DD" format.
- Example: FlightSearch("New York", "London", "2022-10-01") would fetch flights from New York to London on October 1, 2022.

6. BudgetEstimator(origin, destination, dates, people_number)
- Description: Calculate the minimal estimated budget for the trip. Use this tool to verify if the budget provided by the user is sufficient.
- Parameters:
  - origin (str, required): The departure city of your trip.
  - destination (list[str], required): A list of cities you plan to visit during your trip.
  - dates (list[str], required): A list of dates corresponding to the departure from the origin and each of the destinations. The first date is the departure from the origin, and each subsequent date corresponds to the departure from the respective city in the destination list. The last date in this list is the departure from the final destination back to the origin city. All dates should be formatted as "YYYY-MM-DD".
  - people_number (int, required): The number of people on the trip.
- Example: BudgetEstimator("London", ["Paris", "Rome", "Madrid"], ["2022-09-01", "2022-09-05", "2022-09-10", "2022-09-15"], 2) would return the minimal estimated budget for a trip from London to Paris, from Paris to Rome, from Rome to Madrid, and from Madrid back to London on September 1, 5, 10, and 15, 2022, respectively, for two people.
\end{lstlisting}

\end{snippetbox}
\twocolumn
}

{\small
\onecolumn
\begin{snippetbox}[colback=background_blue,colframe=frame_blue,rounded corners]{Static Execution Prompts}{prompt:static_execution}
\textbf{System Message}
\begin{lstlisting}
Based on the provided conversation between the user and the assistant, generate function calls to collect valid information related to accommodations, dining, attractions, transportation, and budget. The available functions are detailed below:

[$\textit{tool description}$]

Please ensure that nested function use is avoided, escape symbols are not included in the string, and functions are only called when all required parameters are available. Your response should include all available function calls, specifying both the function name and its parameters, with each function on a separate line.
\end{lstlisting}
\textbf{Example}
\begin{lstlisting}
Conversation: [{'user': 'Could you create a 3-day travel plan for 7 people from Ithaca to Portland on day 1, from March 8th, 2022?'}, {'assistant': 'Sorry, I couldn\'t find a way to arrive in Portland. Could you provide an alternative destination?'}, {'user': 'Charlotte.'}]

Response:
AccommodationSearch("Charlotte", [])
RestaurantSearch("Charlotte", [])
AttractionSearch("Charlotte")
DistanceMatrix("Ithaca", "Charlotte", "taxi")
DistanceMatrix("Ithaca", "Charlotte", "self-driving")
FlightSearch("Ithaca", "Charlotte", "2022-03-08")
DistanceMatrix("Charlotte", "Ithaca", "taxi")
DistanceMatrix("Charlotte", "Ithaca", "self-driving")
FlightSearch("Charlotte", "Ithaca", "2022-03-10")
BudgetEstimator("Ithaca", ["Charlotte"], ["2022-03-08", "2022-03-10"], 7)
\end{lstlisting}

\textbf{User Message}
\begin{lstlisting}
Conversation: [$\textit{...}$]
Response:
\end{lstlisting}
\end{snippetbox}

\begin{snippetbox}[colback=background_blue,colframe=frame_blue,rounded corners]{Error Prompts in Dynamic Execution}{prompt:dynamic_execution_error}
\textbf{Invalid Action}
\begin{lstlisting}
Invalid action: [$\textit{...}$]. Valid actions include AccommodationSearch(city, filters), RestaurantSearch(city, cuisines), AttractionSearch(city), DistanceMatrix(origin, destination, mode), FlightSearch(origin, destination, date), BudgetEstimator(origin, destination, dates, people_number), and Finish(). Do not include any comment or description in the action.
\end{lstlisting}

\textbf{Invalid Parameters}
\begin{lstlisting}
Invalid parameters for [$\textit{action}$]. Please ensure that all parameters are provided in the correct format.
\end{lstlisting}

\textbf{Null Action}
\begin{lstlisting}
Your action has been filtered due to content restrictions. Please ensure your action does not begin with ['\n', 'Thought', 'Action', 'Observation']. Ensure that the action is permitted in this environment, and try again.
\end{lstlisting}
\end{snippetbox}

\twocolumn
}

{\small
\onecolumn
\begin{snippetbox}[colback=background_blue,colframe=frame_blue,rounded corners]{Dynamic Execution Prompts}{prompt:dynamic_execution}
\textbf{System Message}
\begin{lstlisting}
Based on the provided conversation between the user and the assistant, collect valid information related to accommodations, dining, attractions, transportation, and budget. Solve this task by alternating between "Thought", "Action", and "Observation" steps. "Thought" can reason about the current situation, and "Action" can have 7 different types:

[$\textit{tool description}$]

7. Finish()
- Description: Use this function to indicate the task's completion once all the necessary information has been collected.
- Example: Call Finish() after gathering all necessary information related to accommodations, dining, attractions, transportation, and budget.

Please ensure that nested function use is avoided, escape symbols are not included in the string, and functions are only called when all required parameters are available. Each action should call a single function once, using the valid function name and all required parameters. You should take as many steps as possible until you have gathered the necessary information to complete the task using Finish(). If the user's request is vague or infeasible, avoid making assumptions and strictly use the provided information. Do not add any description or comment to the action. Additionally, do not include line breaks in your response.
\end{lstlisting}

\textbf{User Message for ReAct}
\begin{lstlisting}
Conversation: [$\textit{...}$]
[$\textit{...}$]
\end{lstlisting}
\textbf{User Message for Reflexion}
\begin{lstlisting}
In previous attempts, you tried to use tools to interact with the environment to gather valid information on accommodations, dining, attractions, transportation, and budget given the user-assistant conversation but were unsuccessful. The reflections below offer suggestions to help you avoid past mistakes. Use these insights to refine your strategy for effectively and efficiently utilizing tools to collect the necessary information.
Reflections: 
[$\textit{...}$]
Conversation: [$\textit{...}$]
[$\textit{...}$]
\end{lstlisting}
\textbf{Self-reflection Generation}
\begin{lstlisting}
You are an advanced reasoning agent capable of self-improvement through reflection. You will review a previous attempt where you failed to effectively utilize a tool to gather valid information about accommodations, dining, attractions, transportation, and budget given the user-assistant conversation. Analyze the reasons for the mistake, referencing the tool documentation, the observation, and the action you have taken. Then, formulate a concise, high-level explanation to address and prevent similar errors in the future. Keep your reflections in complete sentences without any line breaks.

Tool documentation:
[$\textit{...}$]
Ensure that each action uses only one non-nested tool and contains no comments or descriptions.

Observation: [$\textit{...}$]
Action: [$\textit{...}$]

Reflection:
\end{lstlisting}
\end{snippetbox}

\twocolumn
}
{\small
\onecolumn
\begin{snippetbox}[colback=background_green,colframe=frame_green,rounded corners]{Static Planning Prompts}{prompt:static_planning}
\textbf{System Message}
\begin{lstlisting}[mathescape=false]
You are a proficient planner tasked with generating a detailed travel plan in JSON format, which is an array of objects, based on the interaction trajectory and the user-assistant conversation. Your plan must strictly adhere to the format provided in the example, incorporating specific details such as flight numbers (e.g., "F0123456"), restaurant names, and accommodation names. Ensure all information in your plan is derived solely from the provided data and aligns with common sense. Attraction visits and meals are expected to be diverse. Use the symbol "-" for unnecessary details, such as "eat at home" or "on the road". For instance, you do not need to plan after returning to the departure city. When traveling to two cities in one day, ensure that "current_city" aligns with the format "from A to B" in the example. If transportation details indicate a journey from one city to another (e.g., from A to B), update the "current_city" to the destination city (in this case, B) the following day. Use ";" to separate different attractions, formatting each as "Name, City". Make sure all flight numbers and costs are appended with a colon (e.g., "Flight Number:" and "Cost:"), consistent with the example. Your JSON plan should include the following fields: ["day", "current_city", "transportation", "breakfast", "attraction", "lunch", "dinner", "accommodation"]. Escape symbols should be used in the string when necessary. Additionally, remove any "$" symbols and comments from the plan.

***** Example *****
Conversation: [{'user': 'Could you create a 3-day travel plan for 7 people from Ithaca to Portland on day 1, from March 8th, 2022?'}, {'assistant': 'Sorry, I couldn't find a way to arrive in Portland. Could you provide an alternative destination?'}, {'user': 'Charlotte.'}, {'assistant': 'It seems you haven't mentioned the expected budget for this trip. Could you provide that information?'}, {'user': 'Yes, my expected budget is $30,200.'}]
Travel Plan: [{"day": 1, "current_city": "from Ithaca to Charlotte", "transportation": "Flight Number: F3633405, from Ithaca to Charlotte, Departure Time: 05:38, Arrival Time: 08:10", "breakfast": "Nagaland's Kitchen, Charlotte", "attraction": "The Charlotte Museum of History, Charlotte;", "lunch": "Cafe Maple Street, Charlotte", "dinner": "Bombay Vada Pav, Charlotte", "accommodation": "Affordable Spacious Refurbished Room in Bushwick!, Charlotte"}, {"day": 2, "current_city": "Charlotte", "transportation": "-", "breakfast": "Olive Tree Cafe, Charlotte", "attraction": "The Mint Museum, Charlotte;Romare Bearden Park, Charlotte;", "lunch": "Birbal Ji Dhaba, Charlotte", "dinner": "Pind Balluchi, Charlotte", "accommodation": "Affordable Spacious Refurbished Room in Bushwick!, Charlotte"}, {"day": 3, "current_city": "from Charlotte to Ithaca", "transportation": "Flight Number: F3786160, from Charlotte to Ithaca, Departure Time: 20:48, Arrival Time: 22:34", "breakfast": "Subway, Charlotte", "attraction": "Books Monument, Charlotte;", "lunch": "Olive Tree Cafe, Charlotte", "dinner": "Kylin Skybar, Charlotte", "accommodation": "-"}]
***** Example Ends *****
\end{lstlisting}

\textbf{User Message for Direct}
\begin{lstlisting}
Interaction trajectory: [$\textit{...}$]
Conversation: [$\textit{...}$]
Travel Plan:
\end{lstlisting}

\textbf{User Message for Zero-shot CoT}
\begin{lstlisting}
Interaction trajectory: [$\textit{...}$]
Conversation: [$\textit{...}$]
Travel Plan: Let's think step by step. First,
\end{lstlisting}
\end{snippetbox}

\begin{snippetbox}[colback=background_green,colframe=frame_green,rounded corners,after=\vspace{0cm}]{Error Prompts in Dynamic Planning}{prompt:dynamic_planning_error}
\textbf{Invalid Plan}
\begin{lstlisting}
The subplan cannot be parsed into JSON format; please check. Only a one-day plan is supported.
\end{lstlisting}

\textbf{Error Plan}
\begin{lstlisting}
The subplan cannot be parsed into JSON format due to the syntax error; please check.
\end{lstlisting}

\textbf{Invalid Action}
\begin{lstlisting}
Invalid action. Valid actions include CostEnquiry(subplan) and Finish(final_json_plan). Please ensure that the parameter is provided in the correct format. Do not include any comments, descriptions, or line breaks in your response.
\end{lstlisting}

\end{snippetbox}

\twocolumn
}

{\small
\onecolumn
\begin{snippetbox}[colback=background_green,colframe=frame_green,rounded corners,after=\vspace{0cm}]{Dynamic Planning Prompts}{prompt:dynamic_planning}
\textbf{System Message}
\begin{lstlisting}[mathescape=false]
You are a proficient planner tasked with generating a detailed travel plan in JSON format, which is an array of objects, based on the interaction trajectory and the user-assistant conversation. Your plan must strictly adhere to the format provided in the example, incorporating specific details such as flight numbers (e.g., "F0123456"), restaurant names, and accommodation names. Ensure all information in your plan is derived solely from the provided data and aligns with common sense. Attraction visits and meals are expected to be diverse. Use the symbol "-" for unnecessary details, such as "eat at home" or "on the road". For instance, you do not need to plan after returning to the departure city. When traveling to two cities in one day, ensure that "current_city" aligns with the format "from A to B" in the example. If transportation details indicate a journey from one city to another (e.g., from A to B), update the "current_city" to the destination city (in this case, B) the following day. Use ";" to separate different attractions, formatting each as "Name, City". Make sure all flight numbers and costs are appended with a colon (e.g., "Flight Number:" and "Cost:"), consistent with the example. Your JSON plan should include the following fields: ["day", "current_city", "transportation", "breakfast", "attraction", "lunch", "dinner", "accommodation"]. Escape symbols should be used in the string when necessary. Additionally, remove any "$" symbols and comments from the plan. Solve this task by alternating between "Thought", "Action", and "Observation" steps. The "Thought" phase involves reasoning about the current situation. The "Action" phase can be of two types:
1. CostEnquiry(subplan): This function calculates the cost of a detailed subplan, for which you need to input the number of people and plan in JSON format. The subplan should encompass a complete one-day plan and include the following fields: ["people_number", "day", "current_city", "transportation", "breakfast", "attraction", "lunch", "dinner", "accommodation"]. An example will be provided for reference.
2. Finish(final_json_plan): Use this function to indicate the completion of the task. You must submit a final, complete plan in JSON as the argument.

***** Example *****
Conversation: [{'user': 'Could you create a 3-day travel plan for 7 people from Ithaca to Portland on day 1, from March 8th, 2022?'}, {'assistant': 'Sorry, I couldn't find a way to arrive in Portland. Could you provide an alternative destination?'}, {'user': 'Charlotte.'}, {'assistant': 'It seems you haven't mentioned the expected budget for this trip. Could you provide that information?'}, {'user': 'Yes, my expected budget is $30,200.'}]
You can call CostEnquiry like CostEnquiry({"people_number": 7, "day": 1, "current_city": "from Ithaca to Charlotte", "transportation": "Flight Number: F3633405, from Ithaca to Charlotte, Departure Time: 05:38, Arrival Time: 08:10", "breakfast": "Nagaland's Kitchen, Charlotte", "attraction": "The Charlotte Museum of History, Charlotte", "lunch": "Cafe Maple Street, Charlotte", "dinner": "Bombay Vada Pav, Charlotte", "accommodation": "Affordable Spacious Refurbished Room in Bushwick!, Charlotte"})
You can call Finish like Finish([{"day": 1, "current_city": "from Ithaca to Charlotte", "transportation": "Flight Number: F3633405, from Ithaca to Charlotte, Departure Time: 05:38, Arrival Time: 08:10", "breakfast": "Nagaland's Kitchen, Charlotte", "attraction": "The Charlotte Museum of History, Charlotte;", "lunch": "Cafe Maple Street, Charlotte", "dinner": "Bombay Vada Pav, Charlotte", "accommodation": "Affordable Spacious Refurbished Room in Bushwick!, Charlotte"}, {"day": 2, "current_city": "Charlotte", "transportation": "-", "breakfast": "Olive Tree Cafe, Charlotte", "attraction": "The Mint Museum, Charlotte;Romare Bearden Park, Charlotte;", "lunch": "Birbal Ji Dhaba, Charlotte", "dinner": "Pind Balluchi, Charlotte", "accommodation": "Affordable Spacious Refurbished Room in Bushwick!, Charlotte"}, {"day": 3, "current_city": "from Charlotte to Ithaca", "transportation": "Flight Number: F3786160, from Charlotte to Ithaca, Departure Time: 20:48, Arrival Time: 22:34", "breakfast": "Subway, Charlotte", "attraction": "Books Monument, Charlotte;", "lunch": "Olive Tree Cafe, Charlotte", "dinner": "Kylin Skybar, Charlotte", "accommodation": "-"}])
***** Example Ends *****

You must use Finish(final_json_plan) to indicate that you have finished the task. Each action only calls one function once, without any comments or descriptions. Do not include line breaks in your response.
\end{lstlisting}

\textbf{User Message for React}
\begin{lstlisting}
Interaction trajectory: [$\textit{...}$]
Conversation: [$\textit{...}$]
[$\textit{...}$]
\end{lstlisting}

\textbf{User Message for Reflexion}
\begin{lstlisting}
You have attempted to give a subplan before and failed. The following reflection(s) give a suggestion to avoid failing to answer the query in the same way you did previously. Use them to improve your strategy for correctly planning.
Reflections: 
[$\textit{...}$]
Interaction trajectory: [$\textit{...}$]
Conversation: [$\textit{...}$]
[$\textit{...}$]
\end{lstlisting}
\end{snippetbox}

\twocolumn
}

\end{document}